\documentclass[mnsc]{informs3_homepage}
\usepackage{natbib,adjustbox,float,comment,booktabs,diagbox,caption,subcaption,graphicx,bm, makecell,textcomp,mathtools,multirow}
\usepackage[table]{xcolor}
\usepackage{hyperref}

\usepackage{enumitem}
\usepackage{bm}
\usepackage{changepage}

\usepackage{hyperref}
\hypersetup{
    colorlinks,
    citecolor=black,
    filecolor=black,
    linkcolor=black,
    urlcolor=black
}
\newcommand{\pkg}[1]{{\fontseries{b}\selectfont #1}}

\newcommand{\NA}{---}

\usepackage{graphicx}
\usepackage[margin=1in]{geometry}  
\usepackage{float}
\usepackage[ruled,vlined]{algorithm2e}
\usepackage{bbm}

\newcommand{\eqs}[1]{\begin{align} #1 \end{align}}
\newcommand{\eqss}[1]{\begin{align*} #1 \end{align*}}

\DeclareBoldMathCommand\balpha{\alpha}
\DeclareBoldMathCommand\bdelta{\delta}
\DeclareBoldMathCommand\bsigma{\sigma}

\newcommand{\vI}{{\mathbf{I}}}

\newcommand{\vX}{{\mathbf{X}}}

\newcommand{\cH}{{\mathcal{H}}}
\newcommand{\cI}{{\mathcal{I}}}

\newcommand{\cL}{{\mathcal{L}}}

\newcommand{\cN}{{\mathcal{N}}}

\newcommand{\one}{\mathbbm{1}}

\newcommand{\RR}{\mathbb{R}}
\newcommand{\JJ}{\mathbb{J}}
\newcommand{\KK}{\mathbb{K}}

\newcommand{\EE}{\mathbb{E}}

\newcommand{\PP}{\mathbb{P}}

\def\*#1{\mathbf{#1}}
\def\+#1{\mathbb{#1}}

\newcommand{\RNum}[1]{\uppercase\expandafter{\romannumeral #1\relax}}

\SetArgSty{textnormal}

\usepackage{tikz}
\usetikzlibrary{decorations.pathreplacing,calc}

\tikzset{brace/.style={decorate, decoration={brace}},
	brace mirrored/.style={decorate, decoration={brace,mirror}},
}

\newcounter{brace}
\newcounter{arrow}
\setcitestyle{open={(},close={)}}

\bibpunct[, ]{(}{)}{,}{a}{}{,}
\def\bibfont{\small}

\TheoremsNumberedThrough
\ECRepeatTheorems
\EquationsNumberedThrough

\begin{document}
	\RUNAUTHOR{}
	\RUNTITLE{Advertising Bandits in High-dimensions}
	\TITLE{
  Advertising Media and Target Audience Optimization via High-dimensional Bandits \vspace{-0.3cm}}
	
	\ARTICLEAUTHORS{
		\AUTHOR{Wenjia Ba}
		\AFF{Graduate School of Business, Stanford University, \EMAIL{wenjiaba@stanford.edu}}
		\AUTHOR{J. Michael Harrison}
		\AFF{Graduate School of Business, Stanford University, \EMAIL{mike.harrison@stanford.edu}}
		\AUTHOR{Harikesh S. Nair}
		\AFF{Graduate School of Business, Stanford University, \EMAIL{harikesh.nair@stanford.edu}}
		\AUTHOR{\vspace{-0.5cm} }
	} 

	\ABSTRACT{We present a data-driven algorithm that advertisers can use to automate their digital ad-campaigns at online publishers. The algorithm enables the advertiser to search across available target audiences and ad-media to find the best possible combination for its campaign via online experimentation. The problem of finding the best audience-ad combination is complicated by a number of distinctive challenges, including (a) a need for active exploration to resolve prior uncertainty and to speed the search for profitable combinations, (b) many combinations to choose from, giving rise to high-dimensional search formulations, and (c) very low success probabilities, typically just a fraction of one percent. Our algorithm (designated LRDL, an acronym for Logistic Regression with Debiased Lasso) addresses these challenges by combining four elemenets: a multiarmed bandit framework for active exploration; a Lasso penalty function to handle high dimensionality; an inbuilt debiasing kernel that handles the regularization bias induced by the Lasso; and a semi-parametric regression model for outcomes that promotes cross-learning across arms. The algorithm is implemented as a Thompson Sampler, and to the best of our knowledge, it is the first that can practically address all of the challenges above. Simulations with real and synthetic data show the method is effective and document its superior performance against several benchmarks from the recent high-dimensional bandit literature.
	}
	\KEYWORDS{\vspace{-0.4cm} Advertising, Marketing, Multi-armed Bandits, Lasso, Regularization, Revenue Management}
	
	\maketitle
	
	\section{Introduction}
The online advertising industry has grown rapidly in recent years,  and a major driver of that growth has been the development of customized ads (called ``creatives'' in this paper) for different audience segments. As advertising targeting has expanded from being based on purely demographic criteria (e.g., age, gender, and geography) to using contextual (e.g., the webpage a user is browsing) and behavioral criteria (e.g., past actions), the number of possible target audience segments the advertiser could choose from to create an effecitve campaign has become enormous. Moreover, there are often many potential creatives, using a variety of different media such as images of various types (e.g., with different ad-content, immersive, full-screen), videos (short or long-form), that could be used for a given target audience. A priori, it is not clear which medium works for which audience and which is a good audience-creative combination for the campaign. The combination of all these factors leads to \textit{high-dimensional} advertising matching problems, by which we mean that the number of potential audience-creative combinations is large relative to the amount of experimentation that an advertiser can conduct. The proliferation of targeting options also increases advertiser uncertainty, and consequently increases the demand for automated experimentation to resolve that uncertainty. This paper is focused on methods for managing and guiding such experimentation so as to help the advertiser deliver the most effective creatives to its potential customers.

There are several obstacles to developing a practical algorithmic solution to the problem. Success rates are low for online ads, with the click-through rate (CTR) on typical ads being about 0.5\%.  To put this in context, in an ad campaign where 1 million impressions are allocated evenly among 1000 combinations of possible audience and creatives, each combination can expect to receive fewer than five clicks (that is, positive responses). Therefore, getting accurate CTR estimates using traditional testing methods, where user traffic is  evenly divided among competing alternatives, may be either impractical or prohibitively expensive in high-dimensional settings. This motivates two central features of the methods discussed in this paper: \textit{adaptive sampling} and \textit{cross learning}. In our context, ``adaptive sampling'' means that as experimentation progresses, sampling is biased toward alternatives that have shown greater promise. ``Cross learning'' means that results from a single audience-ad pair are used, in combination with a statistical model, to make inferences about the effectiveness of other pairs as well.

In this paper, we consider two models of sequential decision-making for online advertising. In the first of them, called the \textit{single-stage model}, decisions are made by the advertiser, or rather by an algorithm that the advertiser provides. For each trial $t = 1, 2, \dots , T,$ the advertiser’s algorithm chooses a \textit{target audience} (TA) for which to purchase an impression, and simultaneously chooses a \textit{creative} (CA) to display for a user chosen at random from that audience segment. The algorithm then observes binary feedback in the form of click or no-click. Building on previous work in the field, the advertiser’s problem is cast as a multiarmed bandit, and more specifically as a \textit{Bernoulli bandit}, with (TA, CA) combinations playing the role of ``arms''. The advertiser aims to make TA and CA choices to maximize the expected click-through rate (a proxy for payoffs) achieved in a total of $T$ trials.

In our second, \textit{two-stage model}, it is the platform that conducts experiments, using an automated algorithm to optimize an ad campaign on behalf of the advertiser. Information plays a more prominent role in this two-stage model than in our single-stage model: Compared to the advertiser, the platform is likely to have additional information it privately observes about arriving impressions, and it can use that finer information in choosing the CA presented to a user, instead of basing the choice on just the broadly defined TA from which the user is drawn. Therefore, in our second model, the problem is treated as a contextual bandit. We compare the click-through rates achieved in these two different learning regimes, characterizing the difference as the benefit that an advertiser realizes by having a platform conduct experiments on its behalf.

The algorithm we present (henceforth referred to as ``LRDL'', for Logistic Regression with Debiased Lasso) is cast in a multiarmed bandit framework that allows for active exploration of various possible audience-ad combinations. A key feature of the method is its use of a semi-parametetric regression model to promote cross-learning across various combinations, with the inclusion of interaction terms to reduce the threat of model misspecification. Specifically, we use a hybrid logistic regression model, which allows parameter sharing between different audience-creative combinations. This enables efficeint learning even in the face of limited observations and low success probabailites per arm. Another key feature of our method is its handling of high dimensionality: to do so, we impose a Lasso penality in each round of model fitting or re-fitting, so that only a small number of interaction terms are used in the model. We regularize the inference objective to ensure that the cross-learning objective remains paramount. As is well known, this induces a regularization bias; to handle this, we adapt the debiasing ideas proposed by \cite{javanmard2014confidence} directly into the bandit inference procedure. Putting it all together, we show how to implement the entire procedure in a coherent way as a generalized Thompson Sampler. To the best of our knowledge, the LRDL method proposed in this paper is the first algorithm that can meet the practical challenges of extreme sparse feedback and high dimensionality in the audience-creative selection problem in online advertising.

To assess the LRDL method, we implement extensive simulations with synthetic data and with real data on ads-campaigns from JD.com, an e-commerce company in China and a large publisher of digital ads. Our simulations show that our proposed method is effective and document its superior performance against several becnhmarks from the recent high-dimensional bandit literature. Our simulations also illuminate several side issues concerning the value, from an advertiser’s perspective, of the platform conducting experiments on the advertiser’s behalf. Agency concerns aside, the advertiser could be better off when a platform conducts experiments on the advertiser’s behalf, because the platform can leverage addiitonal information it possesses. In our simulation studies, we find that this is indeed true when sequential decisions are made using the LRDL method, but perhaps surprisingly, using more detailed information may actually leave the advertiser \textit{worse off} if a naive bandit algorithm is deployed. Roughly speaking, this happens when a simplistic algorithm is overwhelmed by the apparent need to sort through a multitude of decision options. This emphasizes the importance of algorithmic sophistication and the use of principled statistical methods in driving value for both advertisers and platforms.  

The remainder of this introduction briefly characaterizes the relationship of this study to the existing literature and explains its contribution relative to past work. 

\noindent\textbf{Related work.} \quad The problem of how best to target and measure online ads has inspired research in various fields, including machine learning, marketing, and operations. Please see the comprehensive surveys by \cite{wang2017display} on machine learning algorithms for real-time bid ads; \cite{choi2020online} on recent progress in display advertising; and \cite{gordon2021inefficiencies} for experimental approaches to digital advertising measurement.

Within this broader area, the application of online learning methods such as bandits to advertising and digital marketing problems has proliferated in recent years. One branch of this literature studies the specifc problem of bidding in online auctions, cf. \cite{chapelle2015offline, balseiro2015repeated, baardman2019scheduling, balseiro2019learning,waisman2019online,tunuguntla2021near}. Another branch is concerned with the related problem of scheduling different ads, cf. \cite{turner2012planning,hojjat2017unified}. In addition, there are many papers on successful applications of bandit models to web content optimization, including \cite{agarwal2009explore,li2010contextual,chapelle2011empirical, agarwal2014taming, urban2014morphing}, and \cite{agarwal2016making}. 

Within this stream, this paper is most closely related to papers that use bandit models for controlled experiments in online platforms. The works closest to ours are \cite{scott2015multi,schwartz2017customer,ju2019sequential}, and \cite{geng2020online,geng2020comparison}, which propose using bandit experiments to evaluate the effectiveness of different creatives for targeted advertising. In contrast to our work, however, none of those earlier papers have shown how to resolve the tradeoff between exploration and exploitation in a high-dimensional setting, which is our distinctive focus and the main contribution of this paper.

Despite scattered recent progress cited below, the high-dimensional bandit problem remains in many respects an open problem. The recent work in this area falls into two categories. First, the following papers all concern \textit{linear} conextual bandit problems with high-dimensional feature vectors: \cite{abbasi2012online}, \cite{pmlr-v22-carpentier12}, \cite{gilton2017sparse}, \cite{wang2018minimax}, \cite{kim2019doubly}, \cite{bastani2020online}, \cite{hao2020high}. In contrast, our concern here is with modeling Bernoulli trials (that is, binary outcomes), which is a nonlinear problem for which the standard statistical approach is to use a \textit{generalzied} linear model (GLM) like logistic regression. The extension from linear to GLM methods is crucial for our purposes to fit the statisical structure of the data and to improve the efficiency and use of the algorithm. The second category referred to above consists of two recent papers that treat generalzied linear models with high-dimensional feature vectors:  \cite{oh2020sparsity} and \cite{li2021simple}. Each of those works proposes an exploration-free or ``greedy'' method that selects the treatment whose current estimated reward is the highest in each iteration. In their context, the greedy method is justified by a so-called relative symmetry condition, which implies that there is enough randomness in the observed covariates to force an adequate degree of exploration. In our setting, we have no time-varying covariates -- all of our features are indicators. Therefore, we cannot rely on varaitions in covariates as a substitute for exploration as do \cite{oh2020sparsity} and \cite{li2021simple}. The real-data experiment described in Subsection \ref{subsec:real} of this paper testifies to the importance of active exploration in the low CTR setting. To facilitate active exploration in high dimensions, we modify the debiased lasso estimator proposed by \cite{javanmard2014confidence} in a linear setting, making it suitable for nonlinear GLM applications. This extension here has its own unique complexity and novelty, as detailed below.

\noindent\textbf{Debiased Lasso.} \quad An important antecedent of our research is the work by \cite{javanmard2014confidence} on high-dimensional parameter estimation. They describe a method to remove bias in lasso estimates and construct confidence regions for linear regressions. We incorporate a modified version of their method in a multiarmed bandit algorithm (i.e., extending to an an adaptive setting), using the associated confidence regions coming out of the debiased model for active exploration. Our approach is inspired by the working paper of \cite{fan2021sample}, which studies a sampling-based method for high-dimensional linear contextual bandits, also using the result of \cite{javanmard2014confidence}. Our work differs from theirs mainly in two aspects. First, our method is tailored to the challenges described above for online advertising, including binary response and scarce feedback. Second, \cite{fan2021sample} focus on the linear setting, whereas our method addresses a generalized linear model (GLM). The extension from linear to GLM is nontrivial. As noted by \cite{xia2020revisit}, a direct application of \cite{javanmard2014confidence} to a GLM setting may suffer inefficiencies. Because of nonlinearity, the standard sparsity assumption on the information matrix (i.e., that the number of nonzero elements is small) is generally not satisfied, which affects the estimation accuracy of the covariance matrix. Ensuring the accuracy of the covariance matrix is critical, because it drives exploration in the Thompson sampler. We specifically modify the method of \cite{javanmard2014confidence} to address this issue.

 \noindent\textbf{Overview of this paper.} \quad Section \ref{sec:problem} contains a precise formulation of the single-stage problem described above, and Section \ref{sec:method} lays out our proposed solution method. Section \ref{sec:Numerical}  presents the results of numerical experiments on real and synthetic data.

Section \ref{sec:twostage} extends our analysis to the more complicated two-stage problem (the platform's problem) described earlier, which is a type of contextual bandit problem. The comparison with the single-stage model provides insights on whether the advertiser should have the platform conduct experiments on its behalf. In Section \ref{sec:discussion} we discuss some secondary topics related to our proposed method and its performance in numerical experiments. Finally, Section \ref{sec:conclusion} summarizes our paper’s contributions and offers a few concluding remarks.

\section{The advertiser's single-stage problem}  \label{sec:problem}

In a sequence of trials indexed by $t = 1,\dots ,T$, the advertiser can algorithmically choose from a set of target audiences (TAs) provided by the platform\footnote{The platform is motivated to keep the number of TAs available to advertisers manageable, because of the need to operate an impressions market for each TA in the collection. That is, the platform seeks to avoid problems associated with ``thin'' markets. When TAs are defined narrowly, a platform may have difficulty satisfying the demand for impressions, simply because of limited supply. Further, there is a risk that the number of advertisers interested in a narrowly defined TA will be small, leading to weak auction pressure and poor revenue extraction. Also, with a small number of advertisers, the success of such an auction may be overly dependent on the platform’s reserve price, which is difficult to tune given the limited number of transactions occurring for such TAs.} (for example, ``San Francisco users'' or ``male users''), and a set of creatives (CAs) that it has available. Hereafter, TAs are indexed by $k = 1,\dots , K$ and CAs by $r = 1,\dots ,R$.

At the beginning of each trial $t$, the advertiser submits a TA-CA combination $(k_t, r_t)$ to the platform. The platform randomly chooses one user from that TA and presents the creative $r_t$ to that user. After the creative is presented, the user’s reaction $y_t \in \{0, 1\}$ is observed, with 1 corresponding to a click and 0 corresponding to no-click. For the purpose of this paper, we formulate the problem as one of maximizing the expected number of clicks realized over $T$ trials, or equivalently, the expected total click-through rate. One way to think of this objective is that the payoff from the ad is some fixed mark-up of the ad's CTR. This is the approach adopted in many past papers in the literature, abstracting away from two aspects of the real problem, namely, the average cost of impressions for different TAs (which depends on the bidding policy used in the impression market), and the average ``conversion value'' of clicks obtained from different audience segments in response to different ads.\footnote{The focus on CTR as an objective is natural for advertising on e-commerce platforms such as JD.com, cf. \cite{geng2020comparison}: in most cases, the click on the ad will directly land the customer on a product page, thereby generating awareness and visibility for the product. Extending the performance criterion to other metrics such as conversion will be a valuable extension; however, it is non-trivial as conversions is a rarer, fast-changing, and much harder to track outcome. See for instance, \cite{wang2022adaptive} for a recent contribution.} 

We frame the advertiser’s problem as a multi-armed bandit (MAB), with TA-CA pairs playing the role of ``arms,'' and with the objective of maximizing the expected CTR. Let us denote by $p_{kr}$ the probability that a user randomly drawn from target audience $k$ will click in response to creative $r$. From the advertiser's perspective, these probabilities are unknown constants to be learned. To compare alternative policies (see Sections \ref{sec:Numerical}, \ref{sec:twostage} and \ref{sec:discussion}), we take the perspective of an omniscient observer (or oracle) who knows the underlying probabilitries $p_{kr}$, as is standard in the literature. More specifically, a performance measure we consider is the standard quantity called \textit{expected cumulative regret} defined as follows:
\eqs{ R_T&=  T\cdot \pi^* -  \EE\left( \sum_{t=1}^T p_{k_t r_t}\right) \text{where~}  \pi^*= \max_{k,r} ~p_{kr}. \label{eq:regret}
}
Here $\EE(\cdot)$ is an expectation over possible values of the history-dependent and policy-dependent pairs $ (k_1 , r_1 ), \dots , (k_T , r_T )$; operationally, this expectation is computed via Monte Carlo iterations that use the click probabilities $p_{kr}$ as input data.\footnote{This is the procedure used to compute the expected cumulative regret in Subsection \ref{subsec:real}. For synthetically generated test problems, a different but closely related performance measure is used; see Subsection \ref{subsec:synthetic}. } In our context, the expected cumulative regret $R_T$ can be viewed as the expected number of clicks lost, relative to an optimal policy, under the policy embodied in the advertiser’s algorithm, and given the click probabilities $p_{kr}$.

This model deviates from previous work by \cite{geng2020comparison}, in that we directly address the decision problem confronted by an advertiser, instead of the design of a preliminary experiment conducted by the platform on behalf of the advertiser. A second model that we develop in Section \ref{sec:twostage} can be interpreted as the platform’s problem, where the experimenter has more information after a user’s arrival, and has the flexibility to select a creative based on that additional information.

	\section{Proposed LRDL method}  \label{sec:method}

The bandit algorithm that we propose, abbreviated LRDL for Logistic Regression with Debiased Lasso, is designed to address the challenges described earlier: low CTR, a need for cross-learning, and a need for active exploration in high-dimensional problems. It consists of three components. The first is a statistical model (Subsection \ref{2sec:prob}) that specifies click probabilities in terms of more basic parameters. We use a hybrid logistic regression model, which allows parameter sharing between different TA-CA combinations. The second component is a method (Subsection \ref{2sec:estimation}) for estimating model parameters based on past data as it accumulates. To derive the confidence region that is critical for exploration in a high-dimensional setting, we modify the method developed by \cite{javanmard2014confidence} to our generalized linear model. Finally, for treatment selection (that is, choice of an action based on the current parameter estimates), we propose a frequentist version of Thompson sampling, generalized to allow an adjustable degree of emphasis on exploration (see Subsection \ref{2sec:treatment}).

\subsection{Hybrid logistic regression model} \label{2sec:prob}

We assume that the user's feedback $y_{kr}$ is Bernoulli-distributed: $y_{kr} \sim \text{Ber}(p_{kr})$. One modeling approach is to treat the unknown parameters $p_{kr}$ for different $(k, r)$ combinations as independent random variables in a Bayesian framework, with  $p_{kr}\sim\text{Beta} (\alpha_{kr}, \beta_{kr})$. Thanks to the self-conjugacy property of the beta distribution, this approach is computationally convenient and has been successfully implemented by \cite{geng2020comparison}. However, this means that, as model parameters are updated over time, the CTR estimate for each TA-CA pair is based solely on trials involving exactly that pair. In reality, the CTRs for pairs that involve either the same TA or the same CA are likely to be similar, so a method that treats different pairs as independent may be statistically inefficient, especially when the number of combinations gets large.

To exploit the potential commonality between different TA-CA pairs, we deploy a hybrid logistic regression model in which 
\eqs{\label{eq:logistic_model} p_{kr}  = (1+\exp(-c_0 - \alpha_k -\beta_r-\gamma_{kr}))^{-1},}
where $c_0, \alpha_k, \beta_r,\gamma_{kr}$ represents the ``baseline effect'', ``TA effect'', ``CA effect'', and ``additional joint effect'', respectively. Parameters $c_0, \alpha_k,\beta_r$ are shared between different pairs, hence allowing shared information when \textit{combined with suitable estimation methods} (see Subsection \ref{subsec:pos_features}). All parameters ($c_0, \alpha_k, \beta_r, \gamma_{kr} $) are initially unknown to the advertiser and need to be learned through experimentation. In the standard bandit format, this can be expressed as
\eqs{p_{kr} = \exp(\theta^\top \phi(k,r)),}
where 
\eqs{  \label{eq:phi}
\phi(k,r) & = \big(1,  \one_{1}(k),\one_{2}(k), \cdots, \one_{K}(k), \one_{1}(r), \one_{2}(r), \cdots, \one_{R}(r), \one_{1,1}(k,r), \cdots, \one_{K,R}(k,r) \big)^\top,\\ \label{eq:coef}
\theta &= \bigl( c_0, \alpha_1, \cdots,  \alpha_K,  \beta_1, \cdots,\beta_R, \gamma_{11},    \cdots \gamma_{KR} \bigr),
}
and $ \one_{p}(q)$ is an indicator that takes value 1 when $p = q$ and takes value 0 otherwise. That is, the feature vector $\phi(k,r) $ consists of a 1 for the intercept and indicators for TAs, for CAs, and for TA-CA interactions. The unknown parameter vector $\theta$ consists of $c_0,\alpha_k, \beta_{r}$ and $\gamma_{kr}$ values. 

To remove redundant variables in $\theta$, we impose zero-sum constraints $ \sum_k\alpha_k= \sum_r \beta_r=0$ and $\sum_r \gamma_{kr} = \sum_{k} \gamma_{kr}=0$ (for all $k$ and $r$). As suggested in Subsection 4.3 of \cite{buhlmann2011statistics}, the parametrization of the feature vector can be simplified by dropping redundant parameters (e.g., dropping $\alpha_K$, $\beta_R,$ and any $\gamma_{kr} $ that has $K$ or $R$ as part of its subscript).\footnote{This can be automatically implemeted by using the function \pkg{model.matrix} in the \pkg{R} software environment.}  With this assumption, our LRDL method has $d = (K-1) + (R-1) + (K-1)(R-1)+1 = KR $ parameters to estimate, which equals the total number of unknown CTRs that we are trying to estimate. Without further development, then, this statistical model would not actually reduce the number of parameters that need to be estimated. Fortunately, under the sparsity assumption (i..e, that the number of nonzero elements in the unknown parameter $\theta$ is small) that is standard for high-dimensional problems \citep{bastani2020online,oh2020sparsity,li2021simple}, the inferential approach that we use for our  method (debiased lasso) is one that identifies and estimates parameters that are actually influential, setting other parameter values to zero. We show in Subsection \ref{2sec:estimation} that this problem reduction is crucial for cross-learning.

The statistical model described in this subsection is a special case of the \textit{factorization machine model}, which has been used successfully to estimate CTRs in recommendation systems, cf. \cite{rendle2010pairwise} and \cite{ menon2011response}. In contrast with previous applications that focus on estimation and prediction, we use this model in a learning environment; when combined with lasso estimation, it encourages cross-learning and promotes exploration.
    
\subsection{Debiased lasso for parameter estimation} \label{2sec:estimation}

The estimation method we use is adapted from Javanmard and Montanari (2014), which in turn builds on the lasso estimator. The lasso (least absolute shrinkage and selection operator), first proposed by \cite{tibshirani1996regression}, is a popular method for simultaneous estimation and variable selection in high-dimensional models, while retaining computational feasibility for most applications. Unfortunately, this method yields biased estimators, and it is difficult to quantify the uncertainty of estimates (confidence regions) and perform exploration that is crucial for many bandit problems. Until very recently, there were no principled approaches for obtaining confidence regions for high-dimensional parameter vectors, but in 2014 the following three papers proposed related approaches for addressing that problem: \cite{javanmard2014confidence,van2014asymptotically,zhang2014confidence}. In each case the authors construct confidence intervals for lasso estimates by ``debiasing'' the original lasso estimator, that is, adding relatively small terms to the original lasso estimates to yield an unbiased estimator. They show that this debiased lasso estimator has desirable theoretical properties, retains computational feasibility, and performs well empirically. In this paper, we choose the estimator proposed by \cite{javanmard2014confidence}, both for its computational tractability and because it requires the fewest tuning parameters among the three aforementioned estimators.

Because \cite{javanmard2014confidence} developed their method for a standard \textit{linear} statistical model, it must be modified for use with our logistic regression model, which is an example of a \textit{generalized} linear model (GLM). As \cite{xia2020revisit} have observed, a direct extension of \cite{javanmard2014confidence} to a GLM setting may cause bias and an unreliable confidence region, because the key sparsity assumption on the inverse information matrix (that is, the assumption that all but a small number of its elements are zero) may not hold: in the GLM setting, the inverse information matrix depends on the unknown parameter vector, and with the covariates generally correlated, the off-diagonal terms are non-zero. Instead, for the computation of the sample covariance matrix, we adopt the approach in \cite{xia2020revisit} where we directly invert the empirical covariance matrix; this is equation \eqref{eq:dl_cov} in the specification of Algorithm \ref{alg_debiasedLasso} below. 

The debiased lasso estimator builds estimates for $\theta$ (and hence for $p_{kr}$) based on historical data. To start, let $x_\tau = \phi(k_\tau, r_\tau ) \in \RR^d$ be the feature vector and $y_\tau$ be the click or no-click realization associated with trial $\tau$. (Recall that, for each arm or action $(k,r)$ in our model, the corresponding feature vector $\phi(k,r)$ has dimension $d=KR$ after its redundant components are removed.) At time $ t+1$, the data available for the algorithm consists of $\cH_t = \{ (x_\tau, y_\tau) ,\tau = 1,\dots, t\}$. An attractive feature of our statistical model is that its empirical covariance matrix is always invertible. The debiased lasso estimator $ \hat{\theta}_d$ and its corresponding covariance matrix $\hat{\Sigma}_d$ are computed as follows (Algorithm \ref{alg_debiasedLasso}). This algorithm involves a single tuning parameters $c>0$ that is used in setting the penalty rate $\lambda$ for lasso regressions. Our method for determining $c$ will be explained in Section \ref{sec:Numerical}.

\begin{algorithm}
\KwIn{c, history  $\cH_t = \{ (x_\tau, y_\tau), \tau  = 1, \dots, t \}$. \\ 
Set $ d = KR, \lambda = c \sqrt{\log(d)/ d}$.
}
\KwOut{The debiased lasso estimator $\hat{\theta}_d$ and its covariance matrix $\hat{\Sigma}_d $.\\}
Let  $\hat{\theta}_l$ be the lasso estimator calculated by
\eqs{ \label{eq:Lasso}  \hat{\theta}_l  &= \argmin_{\theta} \frac{1}{t} \sum_{\tau=1}^t [ \log \left(1+\exp(x_\tau^\top\theta)\right) - y_{\tau} x_{\tau}^\top \theta  ] + \lambda ||\theta||_1     .}

Compute the empirical covariance matrix \eqs{ \hat{\Sigma} = \frac{1}{t} \sum_{\tau=1}^t  \left( 1+\exp(x_\tau^\top\hat{\theta}_l) \right)^{-1} \left( 1+\exp(-x_\tau^\top\hat{\theta}_l) \right)^{-1} ~ x_{\tau} x_{\tau}^\top.}

Compute the estimator  $\hat{\theta}_d $ and its covariance matrix $  \hat{\Sigma}_d$ as follows:
\eqs{ \label{eq:dl_est}
\hat{\theta}_d &: =  \hat{\theta}_l + \frac{1}{t} \hat{\Sigma}^{-1} \sum_{\tau =1}^t \left(y_\tau -  \left( 1+\exp(-x_\tau^\top\hat{\theta}_l) \right)^{-1} \right) x_\tau,\text{ and~} \\
\label{eq:dl_cov}
  \hat{\Sigma}_d & = \hat{\Sigma}^{-1} / t. }
\caption{Debiased Lasso Estimator for $\theta$ in High-dimensional Logistic Regression Model}
\label{alg_debiasedLasso}
\end{algorithm}

Conceptually, Algorithm \ref{alg_debiasedLasso}  begins by computing the lasso estimator $\hat{\theta}_l$ from equation \eqref{eq:Lasso}. It then obtains an unbiased estimator $\hat{\theta}_d $ by adding a (small) correction, $ \frac{1}{t} \hat{\Sigma}^{-1} \sum_{\tau =1}^t \left(y_\tau -  \left( 1+\exp(-x_\tau^\top\hat{\theta}_l) \right)^{-1} \right) x_\tau$, to $\hat{\theta}_l$.  The main idea of this debiasing approach is to invert the Karush-Kuhn-Tucker (KKT) characterization of the lasso. It is well-known that the lasso estimator $\hat{\theta}_l$ meets the KKT conditions:
\eqs{\label{eq:kkt1}
\frac{1}{n} \vX^\top (Y - \vX \hat{\theta}_l) = \lambda v(\hat{\theta}_l)
}
where $v(\hat{\theta}_l) \in \RR^p$ is a vector in the subgradient of the $\ell_2$ norm at $\hat{\theta}_l$. Plugging into $Y = \vX \theta_0 + \epsilon,$  we have
\eqss{
(\vX^\top \vX/n) (\hat{\theta}_l - \theta_0) +\lambda v(\hat{\theta}_l) = \vX^\top \epsilon/n.
}
Denoting by $\hat{\Theta}$ the ``inverse'' of $\vX^\top \vX/n$, and combining this with \eqref{eq:kkt1}, we then have
\eqs{ \label{eq:kkt2}
\theta_0 = \hat{\theta}_l + \hat{\Theta}\lambda \frac{1}{n} \vX^\top (Y - \vX \hat{\theta}_l).
}
 We omit further discussion of this estimator, referring readers to \cite{van2014asymptotically}, \cite{javanmard2014confidence}, and \cite{zhang2014confidence}. Heuristically, the true parameter vector $\theta$ has a distribution similar to $\cN (\hat{\theta}_d, \hat{\Sigma}_d )$.

\subsection{Generalized Thompson sampling for treatment selection} \label{2sec:treatment}

Hereafter, the letters TS will be used as an abbreviation for ``Thompson Sampling.'' In the classical TS literature (see, for example, \cite{russo2017tutorial}), at each step, the algorithm samples a parameter vector $\tilde{\theta}$ from the current posterior distribution, then selects an action that maximizes the expected reward assuming the true parameter vector equals the sampled value $\tilde{\theta}$. In LRDL, we modify classical TS in two ways. First, because calculating and sampling from a Bayesian posterior can be complicated in a high-dimensional setting, we use as our ``posterior distribution'' for $\theta$ the Gaussian distribution $\cN(\hat{\theta}_d, \hat{\Sigma}_d )$, defined in \eqref{eq:dl_est} and \eqref{eq:dl_cov}, from which the debiased lasso computes confidence regions for $\theta$. Second, we sample $\tilde{\theta}$ from the modified Gaussian distribution $\cN(\hat{\theta}_d, \rho^2\hat{\Sigma}_d )$, viewing the hyper-parameter $\rho$ as a means of adjusting or controlling the rate of exploration: when $\rho=0$, LRDL reduces to a greedy method, and when $\rho=1$, it is the standard form of TS (that is, we directly sample from the ``posterior distribution'' of $\theta$.) The adjustable exploration idea has also been used by \cite{min2020policy} and \cite{kveton2020randomized}. How to find the best exploration index $\rho$ is an interesting research problem in its own right, and is beyond the scope of this paper. Instead, we provide numerical studies in Subsection \ref{subsec:adjustable_exp} on the relationship between the best $\rho$ and problem dimension $d$.

After sampling $\Tilde{\theta}$ from $  \cN(\hat{\theta}_d, \rho^2\hat{\Sigma}_d )$, the algorithm computes
\eqss{
\tilde{p}_{kr} & = \left(1+ \exp\left(-\Tilde{\theta}^\top \phi(k,r) \right)\right)^{-1}}
and sets
\eqss{(k_t, r_t) & = \argmax_{k, r} ~\tilde{p}_{kr}
.}

The various steps that constitute our LRDL method are summarized as Algorithm \ref{alg_LRDL} below. Appendix \ref{app:calc} provides further details on the calculations that underlie Algorithm \ref{alg_LRDL}.
\begin{algorithm}[H]
\KwIn{ $ c, \rho, \hat{\theta}_d  = [0, 0, \cdots, 0] \in \RR^d, \hat{\Sigma}_d = \vI_d$\\
Set  $d = KR, \lambda = c \sqrt{\log d/ t}$.
}

\For{$t=1,2, \cdots, T$}{
Sample $ \Tilde{\theta} \sim \cN(  \theta_d , \rho^2 \Sigma_d)$ \\
Select $ (k_t, r_t) = \argmax_{k, r}  \left(1+ \exp\left(-\Tilde{\theta}^\top \phi(k,r) \right)\right)^{-1} $.\\
Platform sample one user from TA $k_t$ and display CA $r_t$, observe $y_t$.\\
Update $ \cH_t = \cH_{t-1}\cup \{(x_t, y_t)\}$, where $x_t = \phi(k_t,r_t)$.\\  
Compute  $\hat{\theta}_l  = \argmin_{\theta} \frac{1}{t} \sum_{\tau=1}^t [ \log \left(1+\exp(x_\tau^\top\theta)\right) - y_{\tau} x_{\tau}^\top \theta  ] + \lambda ||\theta||_1$, and
$ \hat{\Sigma} = \frac{1}{t} \sum_{\tau=1}^t  \left( 1+\exp(x_\tau^\top\hat{\theta}_l) \right)^{-1} \left( 1+\exp(-x_\tau^\top\hat{\theta}_l) \right)^{-1} \cdot x_{\tau} x_{\tau}^\top$.

Update $\hat{\theta}_d  \leftarrow  \hat{\theta}_l + \frac{1}{t} \hat{\Sigma}^{-1} \sum_{\tau =1}^t \left(y_\tau -  \left( 1+\exp(-x_\tau^\top\hat{\theta}_l) \right)^{-1} \right) x_\tau,$ and $ \hat{\Sigma}_d \leftarrow \hat{\Sigma}^{-1} / t.$
}
\caption{LRDL Algorithm}
 \label{alg_LRDL}
\end{algorithm}

\subsection{Positive features of LRDL } \label{subsec:pos_features}

\noindent \textbf{Cross-learning}. \quad The combination of high-dimensionality, very low click-through rates, and a limited learning horizon ($T$) make it important for the advertiser to use indirect evidence for inference. To understand how LRDL facilitates such cross-learning, first consider a two-way additive logistic regression model $ p_{kr} = (1+e^{-c_0-\alpha_k- \beta_r})^{-1}$, where all parameters ($c_0, \alpha_k, \beta_r$) are shared among different arms. Despite the high potential of this model for cross-learning, there is also a high potential for model misspecification. That is, it is likely that no model of this simple form can fit the true parameters closely.

In response to that concern, we consider adding interaction terms, that is, fitting a model of the form $ p_{kr} = (1+e^{-c_0-\alpha_k -\beta_r -\gamma_{kr}})^{-1}$. This is referred to as a hybrid bandit model by \cite{chapelle2011empirical}, because some of the parameters ($c_0, \alpha_k, \beta_r$) are shared among arms, whereas each of the interaction terms $\gamma_{kr}$ is unique to one arm. Introducing interaction terms reduces the potential for model misspecification, but it may weaken cross-learning if the model is used with an unpenalized estimator. For example, let us denote by $\Tilde{p}_{kr} = \frac{s_{kr}}{n_{kr}}$  the observed success rate in some number of initial trials, where $s_{kr}$ and $n_{kr}$ are the cumulative clicks and number of trials for the TA $k$ and CA $r$ combination. An unpenalized estimation method can achieve a perfect fit to that data by taking $\gamma_{kr} = \log \frac{\Tilde{p}_{kr}}{1-\Tilde{p}_{kr}} $ and $c_0=\alpha_k = \beta_r = 0$ for all $k$ and $r,$ but those parameter estimates make no use at all of indirect evidence. The debiased lasso estimation method addresses this tradeoff by restricting the number of interaction terms (through lasso), which promotes cross-learning, while still constructing an unbiased estimator (through debiasing).

\noindent \textbf{Adaptability}. \quad For Thompson sampling algorithms, the idea is to first sample a parameter estimate from the posterior distribution (or, as in our setting, from the Gaussian distribution that approximates the distribution of the debiased lasso estimator), and then treat it as the actual parameter for decision-making. Once we have sampled a parameter vector, treatment selection is straightforward. Therefore, Thompson sampling algorithms (hence LRDL) can be easily adjusted to complicated settings, as we will show in Section \ref{sec:twostage}. In contrast, another popular class of bandit algorithms, the upper confidence bound algorithms (UCB, for example), involve calculating an optimistic estimate (upper confidence bound) of the expected reward for each arm. The calculation of the optimistic estimate can be very sensitive to problem complexity, so UCB algorithms are usually applied to problems with a simple structure.

	\section{Numerical experiments with the single-stage model  }\label{sec:Numerical}

In this section, we evaluate the performance of our LRDL method against that of existing algorithms, first using real data collected by \cite{geng2020comparison} (Subsection \ref{subsec:real}), and then on synthetically generated test problems (Subsection \ref{subsec:synthetic}). In the first of these  studies, we also explore the sensitivity of LRDL performance to the hyper-parameter $c$ in the formula $\lambda_t = c \sqrt{\log d/t}.$

\subsection{Real Data from JD.com} \label{subsec:real}

For our current purposes, a ``test problem'' is simply a $K\times R$ matrix of numbers $p_{kr} \in (0, 1)$ that represent the true click probabilities (initially unknown, to be learned through experimentation) for the various (TA,CA) combinations $(k, r)$. As stated earlier, such a problem is said to have dimension $d = KR$.

The test problem considered here has dimension 104, featuring 4 TAs and 26 CAs. The complete matrix of click probabilities is displayed in Appendix \ref{app:real} (see Table \ref{tb:realCTR}), where we also explain the process by which this problem was constructed using real data collected from the online platform \hyperlink{https://corporate.jd.com/}{JD.com}, China’s second-largest e-commerce platform.

In our numerical experiments, we perform batch updates with batch size $100$. In other words, each algorithm will generate i.i.d samples $\tilde{\theta}$ from the same distribution for all trials within a batch, and only update the parameter estimates at the end of each batch.

\noindent \textbf{Benchmarks.} \quad Problems of this size occupy an intermediate zone between unambiguously ``high-dimensional'' and unambiguously ``low-dimensional,'' but the former designation is the more reasonable one in practice, as follows. With 100 or more TA-CA combinations under consideration in a bandit experiment, applying a standard bandit algorithm like GLM-UCB or LMLA (see Subsection \ref{subsubsec:low} below) will consume more computing power than is tolerable in everyday practice. For this reason we compare our LMDL algorithm only against two existing methods that are computationally feasible in the high-dimensional domain, namely, the DBBM method used by \cite{geng2020comparison}  and the sparsity agnostic lasso (SA Lasso) method propounded by \cite{oh2020sparsity}. The DBBM method assumes $p_{kr} \sim \text{Beta}(\alpha_{kr},\beta_{kr})$ initially, and it updates those beta distributions independently as data accumulates, so updates involve only addition and subtraction. The SA Lasso method \citep{oh2020sparsity} directly uses lasso parameter estimates for treatment selection (thus it is a ``greedy'' bandit algorithm), and we combine it with logistic regression, using the same feature vector described in \eqref{eq:phi}. We use \cite{oh2020sparsity} as a benchmark because the authors of that paper show that SA Lasso consistently outperforms other existing high-dimensional bandit algorithms. 

\begin{figure}[H]
    \centering  \includegraphics[width=0.45\textwidth]{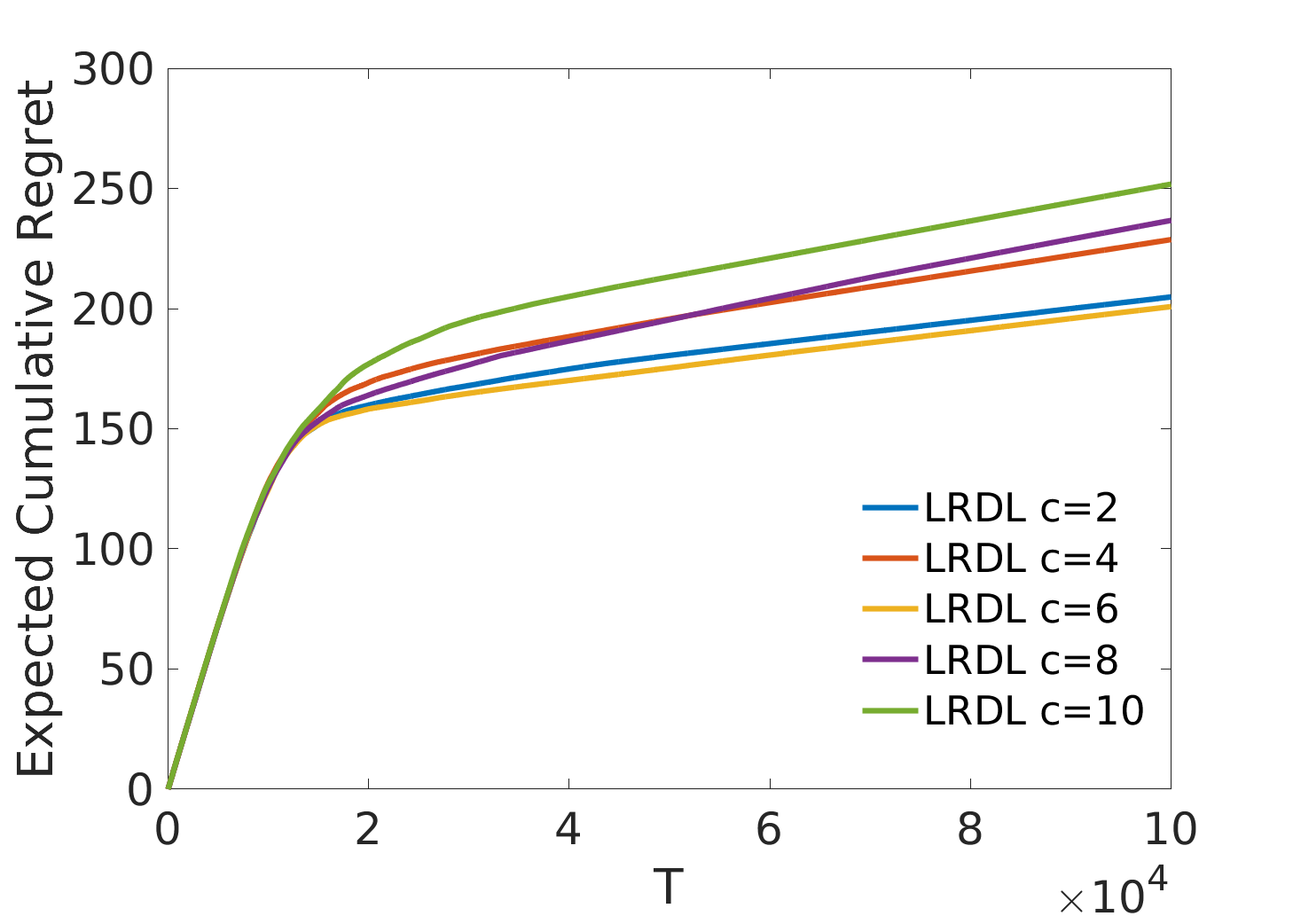}
\linebreak
    \caption{
    Expected cumulative regret for LRDL with different tuning parameter $c \in \{2,4,6,8,10\}$ and $\rho = 0.1$. The displayed statistics are averaged over 40 Monte Carlo replications. We observe that LRDL outperforms both DBBM and SA Lasso.
    }
    \label{fig:real_tuning}
\end{figure}

\noindent\textbf{Algorithm inputs.} \quad The ``hyperparameters'' (tuning parameters) in bandit algorithms are often unknown in practice, e.g., $\alpha$ for SA Lasso \citep{oh2020sparsity}, and $c $ and $\rho$ for LRDL (Algorithm \ref{alg_LRDL}). For each method being compared, we tuned the value of its hyperparameter in the appropriate range to roughly find the best input that minimizes the expected cumulative regret. This led to $\alpha = 0.02$ for SA Lasso \citep{oh2020sparsity}, where we searched in the range $[0.01, 10]$ with precision 0.01. We set  $c = 6$ and $\rho = 0.1$ for LRDL (Algorithm \ref{alg_LRDL}) based on results from Figure \ref{fig:real_tuning} and Figure \ref{fig:explore_real}. 

\begin{figure}[H]
    \centering  \includegraphics[width=0.45\textwidth]{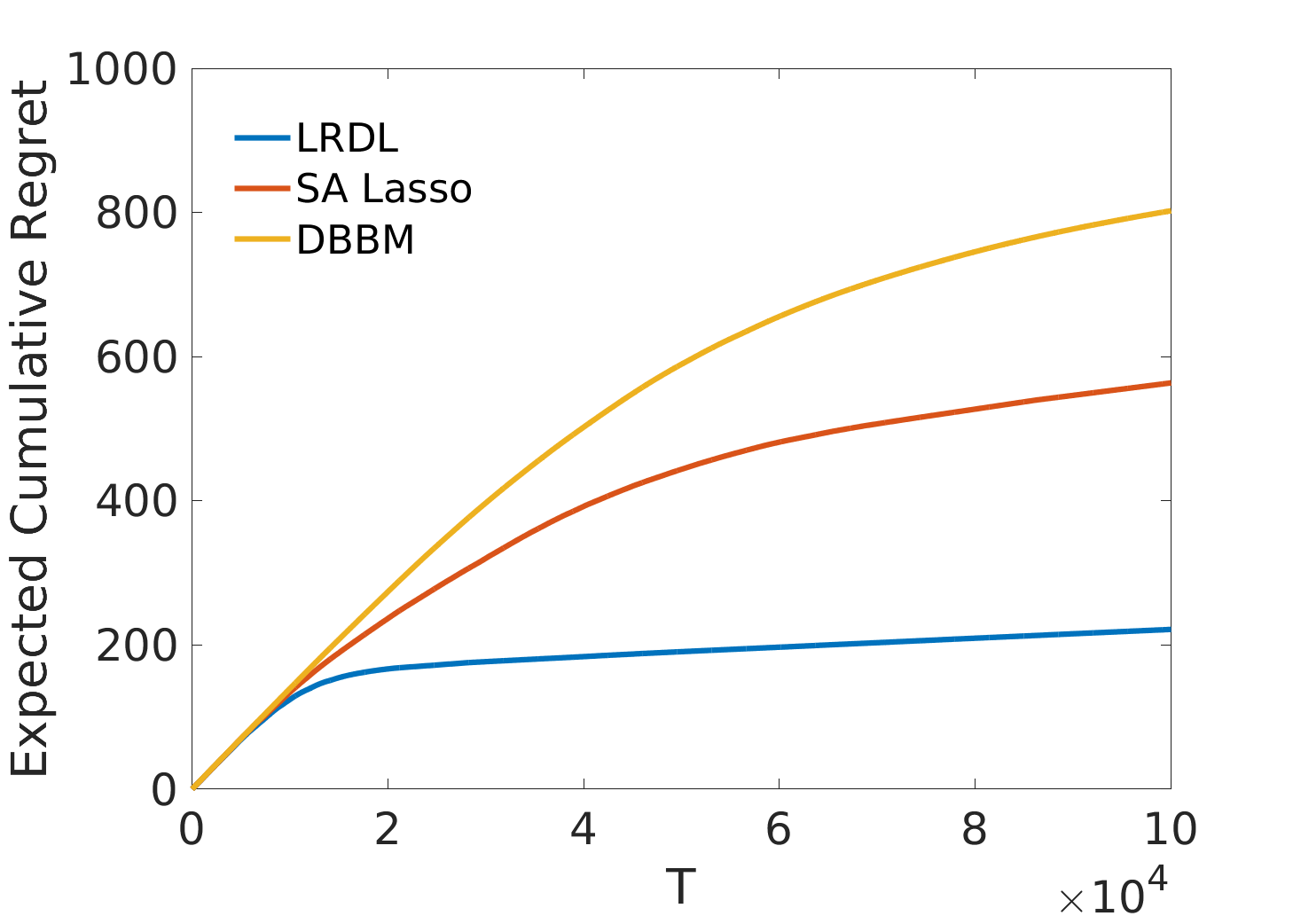}
\linebreak
    \caption{
    Expected cumulative regret for LRDL (Algorithm \ref{alg_LRDL}), DBBM \citep{geng2020comparison}, and SA Lasso \citep{oh2020sparsity} for real data ($ K = 4, R=26$). The displayed statistics are averaged over 40 Monte Carlo replications. We observe that LRDL outperforms both DBBM and SA Lasso.
    }
    \label{fig:real}
\end{figure}

\noindent \textbf{Results.} \quad Below, Figure \ref{fig:real} compares the expected cumulative regret of LRDL, defined as in \eqref{eq:regret}, with that of SA Lasso and DBBM for $ T = 100,000$ trials. Because we attribute a reward of 1 to each click, and a reward of zero to no-click, one may say that regret is expressed in clicks foregone. The number plotted is averaged over 40 Monte Carlo replications. As indicated in Figure \ref{fig:real}, the LRDL method has superior regret performance over both DBBM and SA Lasso. DBBM (yellow line) learns very slowly over 100,000 trials, because it learns CTRs for different TA-CA combinations independently.

SA Lasso (red line) has improved performance over DBBM by inducing cross-learning through hybrid logistic regression and lasso. However, it is still outperformed by LRDL. This is because SA Lasso bases its exploration only on the variability coming from the context (see the relative symmetry assumption in \cite{oh2020sparsity}). As our feature vector is binary with an initial component of 1 (fixed) corresponding to the intercept term $c_0$, there isn't sufficient randomness from the context, so SA Lasso suffers from lack of exploration.

Table \ref{tb:real} presents essentially the same information in tabular form, but with performance expressed in terms of click-through rate rather than regret. We see that LRDL significantly outperforms DBBM and SA Lasso, especially over intermediate horizon lengths ($T$ = 10k, 50k). LRDL achieves 96\% of the maximum possible reward over the first 50,000 trials, even though the environment generates fewer than 500 clicks for more than 100 TA-CA combinations.

\begin{table}[H]
\caption{Reward comparisons of LRDL (Algorithm \ref{alg_LRDL}), DBBM \citep{geng2020comparison}, and SA Lasso \citep{oh2020sparsity} for real data ($K = 4, R=26$). Figures in the body of table show CTR in absolute terms (clicks per trial), and then as percent of maximum possible.}
\smallskip
\label{tb:real}
\centering
\begin{tabular}{l|ll|ll|ll|ll} \hline
  & \multicolumn{2}{l|}{T = 5k} & \multicolumn{2}{l|}{T = 10k} & \multicolumn{2}{l|}{T = 50k} & \multicolumn{2}{l}{T = 100k} \\ \hline
LRDL     & 0.0031       & 19.2\%      & 0.0062       & 38.7\%       & 0.0154       & 96.0\%       & 0.0154        & 96.2\%       \\
SA Lasso     & 0.0027       & 16.8\%      & 0.0041       & 25.6\%       & 0.0116       & 72.5\%       & 0.0143        & 89.2\%       \\
DBBM & 0.0021       & 13.0\%      & 0.0024       & 14.8\%       & 0.0086       & 53.6\%       & 0.0137        & 85.6\%      \\ \hline
\end{tabular}
\end{table}

Figure \ref{fig:real_tuning} shows the sensitivity of LRDL to the tuning parameter $c$. For the experiment, we set $\rho=0.1$ and varied $c \in \{2, 4, 6, 8, 10\}$. Overall, the expected cumulative regret lies in the range of $[200, 250]$ and remains stable despite changes in tuning parameters.

In summary, our experiment shows the importance of cross-learning and active exploration through simulations on real data. LRDL is shown to reduce regret by at least 60\% compared with the second best (SA Lasso), and it is able to achieve about 98.6\% of optimal reward within the first 50,000 trials, where the total clicks are fewer than 500 over all 104 combinations. In addition, LRDL is not very sensitive to the tuning parameter $c.$

\subsection{Synthetically generated test problems} \label{subsec:synthetic}

We evaluate LRDL on synthetically generated data sets to understand how its performance is affected by environmental parameters. In our synthetic test problems, we consider both small (4 TA, 7 CA) and large ones (10 TA, 100 CA), where the true CTR values are generated using the following formula: 
\eqs{ \label{eq:pkr}  p_{kr} = (1+\exp(-c_0-\kappa\alpha_k-  \kappa\beta_{r} - \gamma_{kr}))^{-1}.} 
That is, we generate the true CTR values using a hybrid logistic model of the same form assumed in our LRDL method. (Of course, the LRDL method does not know the true parameter values initially.) The parameters $c_0, \alpha_k, \beta_r,$ and $\gamma_{kr}$ are referred to as the \textit{constant term}, a \textit{TA effect}, a \textit{CA effect}, and an \textit{interaction effect}, respectively. The multiplier $\kappa$ is applied to \textit{both the TA effect and the CA effect}, where the only values we consider in our synthetic test problems are $\kappa=0$ and $\kappa=1$. In the former case there are no TA or CA effects present in the true CTR values, and hence no opportunity for cross-learning, but in the latter case the TA and CA effects are pronounced. 

For each comparison, we are interested in a variant of the expected cumulative regret defined earlier via \eqref{eq:regret}. In particular, let us index by $l = 1,\cdots, L$ the various synthetic test problems and define
\eqss{
\Bar{R}_T^l = T \pi^{*,l} - \sum_{t=1}^T p_{k_tr_t}^l, \text{where~} \pi^{*,l} = \max_{k,r} p_{k_t r_t}^l 
}
which one may describe as a \textit{problem-specific and path-dependent pseudo regret}. In this context, we define the average cumulative regret under a given policy (like GLM-UCB or LRDL) as 
\eqss{
R_T = \frac{1}{L} \sum_{l=1}^L \Bar{R}_T^l.
}
Our adoption of this performance measure is motivated by two considerations. On the one hand, we average over many randomly generated test problems (that is, over many randomly generated parameter combinations) in order to avoid potential reliance on extreme examples whose results are not truly representative. On the other hand, we use a single sample path of the choice sequence $(k_1, r_1), \cdots, (k_T, r_T)$ for each test problem $l$, rather than calculating \textit{expected} cumulative regret for each test problem via Monte Carlo replication as in \eqref{eq:regret}, because the latter procedure would impose an excessive computational burden.

\subsubsection{Low-dimensional problems} \label{subsubsec:low} We consider $K = 4$ and $R  = 7$. For each $\kappa \in \{0, 1\}$, we randomly generate 100 test problems using the procedure explained below, and for each algorithm considered, we report the average cumulative regret over those 100 problems. The CTRs of each test problem are generated via \eqref{eq:pkr}, where only a randomly chosen subset of covariates ($\alpha, \beta,\gamma$) are set to nonzero values. Specifically, we randomly select two $\alpha_k$ components, two $\beta_r$ components, and four $\gamma_{kr}$ components to be nonzero, and draw their values independently from a uniform distribution on [-1,1]. 

\noindent \textbf{Benchmark.} \quad In the low-dimensional setting, we compare LRDL against (a) the Disjoint Beta Bernoulli method (DBBM) employed by \cite{geng2020comparison}, (b) the Logit Model with Laplace Approximation (LMLA) method used by \cite{chapelle2011empirical}, and (c) the GLM-UCB method of \cite{li2017provably}. The LMLA method approximates the posterior distribution by a Gaussian distribution and samples from this simpler distribution. GLM-UCB constructs upper confidence bounds for CTR estimates and uses them for treatment selection. For both LMLA and GLM-UCB, we employ logistic regression with the same feature vector described in \eqref{eq:phi}. 

\noindent\textbf{Algorithm inputs.} \quad For each method being compared, we tuned the value of its hyperparameter in the range $[0.01, 5]$ to roughly find the best input that minimizes regret. This led to the value $\alpha=1$ for GLM-UCB and the values $c = 4$ and $ \rho = 0.1$ for LRDL.
 
\begin{figure}[H]
    \centering
    \subfloat[$\kappa = 1$ ]{\includegraphics[width=0.45\textwidth]{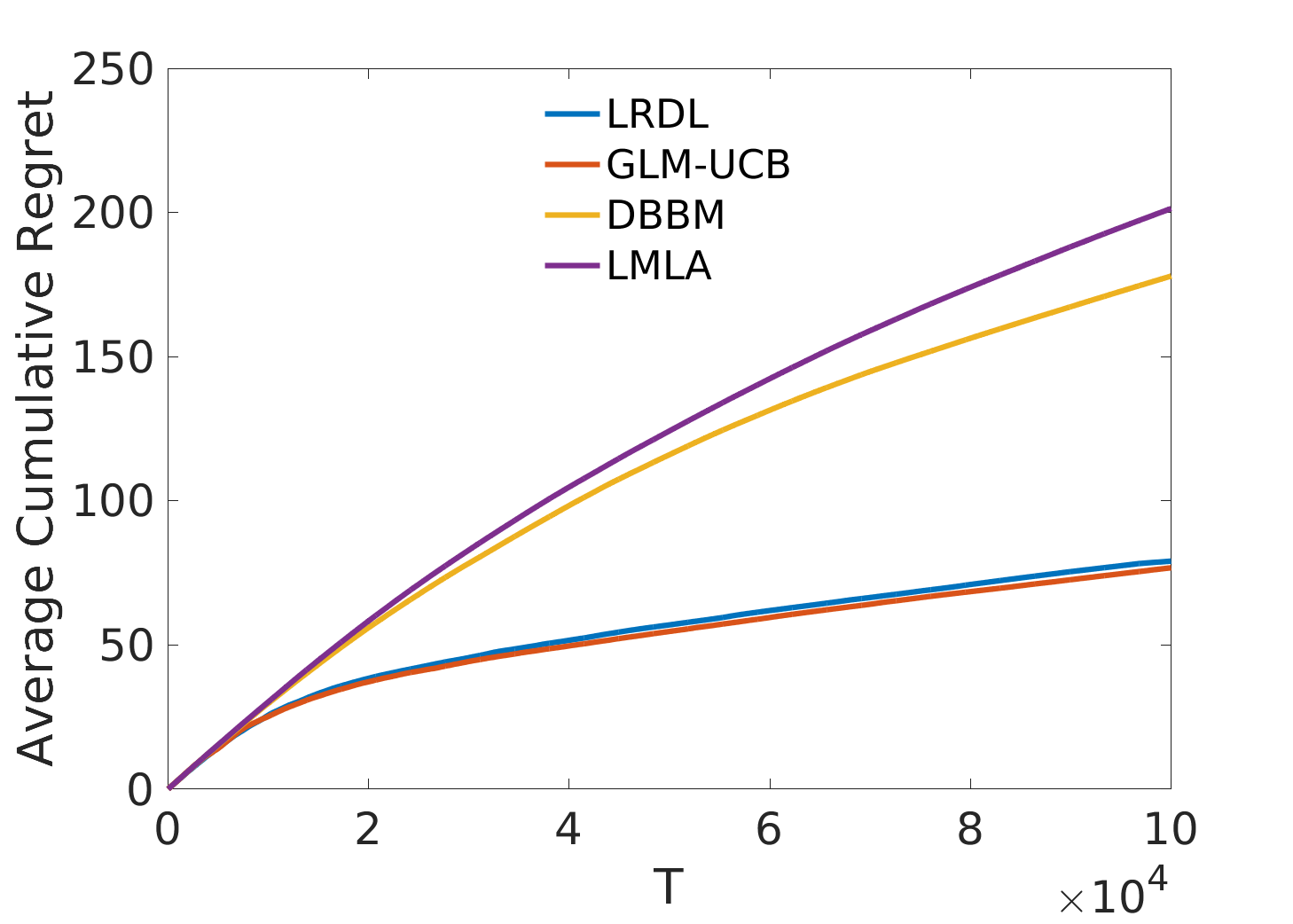}}
\subfloat[$\kappa = 0$]{\includegraphics[width=0.45\textwidth]{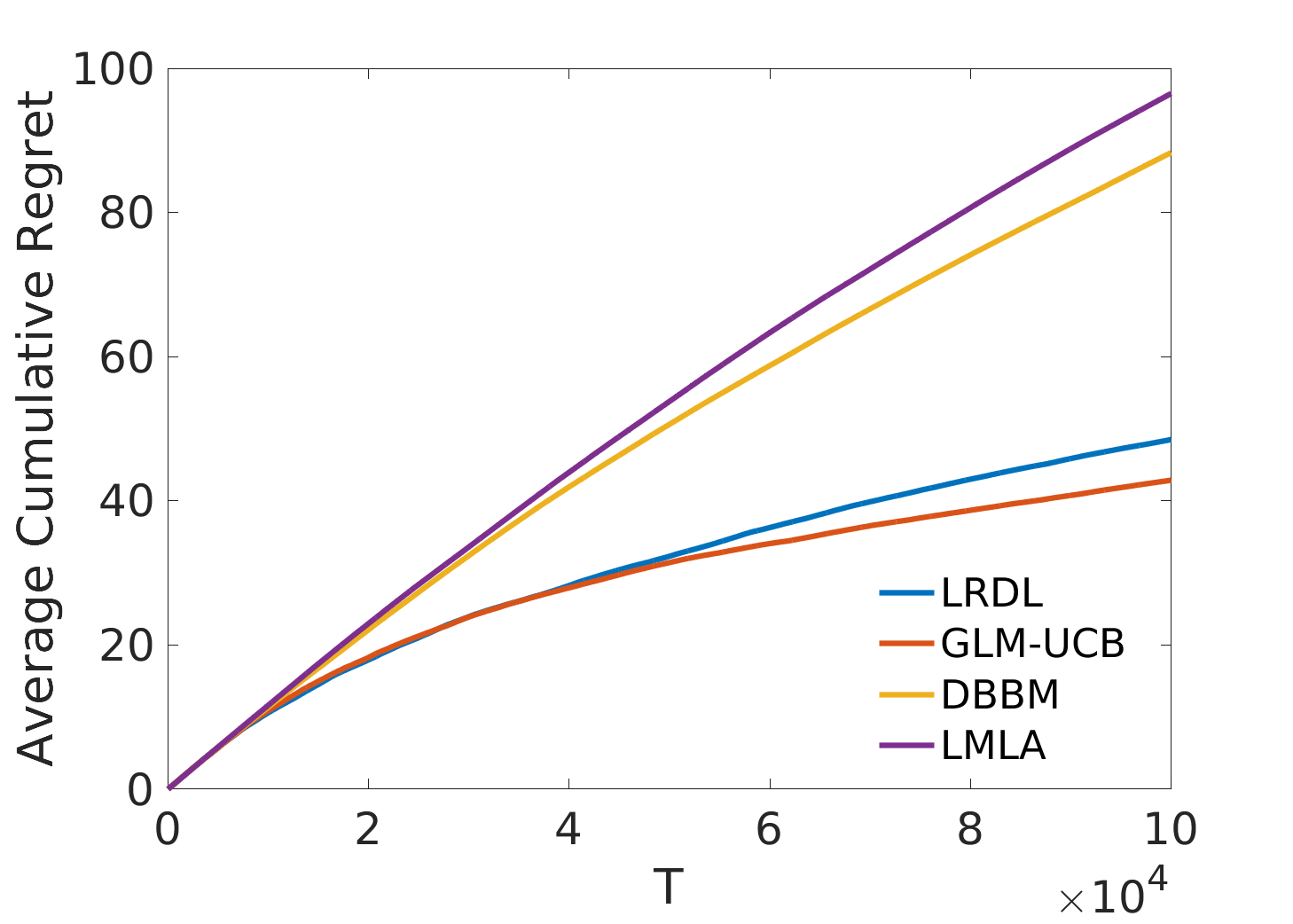}}
\linebreak 
\caption{Cumulative regret of LRDL (Algorithm \ref{alg_LRDL}), DBBM \citep{geng2020comparison}, LMLA \citep{li2010contextual}, and GLM-UCB \citep{li2017provably} for low-dimensional synthetic problems ($K=4, R=7$). We consider the setting with strong CA and TA effect ($\kappa=1$) and with no CA and TA effect ($\kappa=0$). The displayed statistics are averaged over 100 independently generated test problems. We observe that LRDL is competitive (with regard to regret) against GLM-UCB, and significantly outperforms LMLA and DBBM.  }
    \label{fig:small_regret}
\end{figure}

\noindent\textbf{Results.} \quad We first present the average cumulative regret of each algorithm in Figure \ref{fig:small_regret} for both $\kappa = 0$ and $\kappa = 1$.  We observe that LRDL is competitive against GLM-UCB, and clearly outperforms LMLA and DBBM. Next, Table \ref{tb:low} shows both the average reward and the reward ratio. We see that the LRDL and GLM-UCB methods have much better performance in the early stage of the experiment ($T$=5k, 10k), both achieving more than 80\% of maximum reward over the first 10,000 trials even though the environment generates less than 140 clicks. 

\begin{table}
\caption{Reward comparisons of LRDL (Algorithm \ref{alg_LRDL}), DBBM \citep{geng2020comparison}, LMLA \citep{li2010contextual}, and GLM-UCB \citep{li2017provably} for low-dimensional synthetic problems ($K=7, R=4$). Figures in the body of table show CTR in absolute terms (clicks per trial), and then as percent of maximum possible. }
\smallskip
\centering{
\begin{tabular}{l|l|ll|ll|ll|ll} \hline
\multicolumn{2}{l|}{ } & \multicolumn{2}{l|}{T = 5k} & \multicolumn{2}{l|}{T = 10k} & \multicolumn{2}{l|}{T = 50k} & \multicolumn{2}{l}{T = 100k} \\ \hline
\multirow{4}{*}{$\kappa = 0$}           & LRDL             & 0.0077       & 67.4\%      & 0.0096       & 82.1\%       & 0.0112       & 96.0\%       & 0.0113        & 96.9\%       \\
                                     & DBBM             & 0.0052       & 50.3\%      & 0.006        & 55.6\%       & 0.01         & 85.0\%       & 0.0108        & 92.0\%       \\
                                     & LMLA             & 0.0047       & 47.2\%      & 0.0051       & 50.0\%       & 0.0092       & 78.8\%       & 0.0103        & 88.1\%       \\
                                     & GLM UCB          & 0.0091       & 77.8\%      & 0.0103       & 85.0\%       & 0.0112       & 95.1\%       & 0.0112        & 95.1\%       \\ \hline
\multirow{4}{*}{$\kappa = 1$}           & LRDL              & 0.0108       & 67.0\%      & 0.0138       & 85.6\%       & 0.0152       & 98.6\%       & 0.0151        & 98.9\%       \\
                                     & DBBM             & 0.0067       & 46.6\%      & 0.0095       & 60.1\%       & 0.0138       & 87.2\%       & 0.0147        & 93.8\%       \\
                                     & LMLA             & 0.0057       & 42.1\%      & 0.0073       & 49.9\%       & 0.0133       & 82.9\%       & 0.0143        & 90.8\%       \\
                                     & GLM UCB          & 0.0124       & 75.3\%      & 0.0141       & 88.2\%       & 0.015        & 97.1\%       & 0.0151        & 97.6\%   \\ \hline    
\end{tabular}}
\linebreak 
\label{tb:low}
\end{table}

\subsubsection{High-dimensional problems}\label{subsec:highdimResult} We consider $K = 10$ and $R = 100$. For each $\kappa \in \{0, 1\}$, we generate 40 test problems and report the average of the cumulative regret for each algorithm. The CTRs of each test problem are generated via \eqref{eq:pkr}. Again, all parameters ($ \alpha_k, \beta_r, \gamma_{kr}$) are set to 0 except a randomly selected subset. We randomly select five $\alpha_k $ components, five $\beta_r$ components and twenty $\gamma_{kr}$ components to be nonzero, and draw their values independently from a uniform distribution on [-1,1].

\noindent\textbf{Algorithm inputs.} \quad For SA Lasso \citep{oh2020sparsity}, we set $\alpha = 0.02$. For LRDL, we set $ c = 4$ and $ \rho = 0.1$. 

\begin{figure}[H]
    \centering
    \subfloat[$\kappa = 1$]{\includegraphics[width=0.45\textwidth]{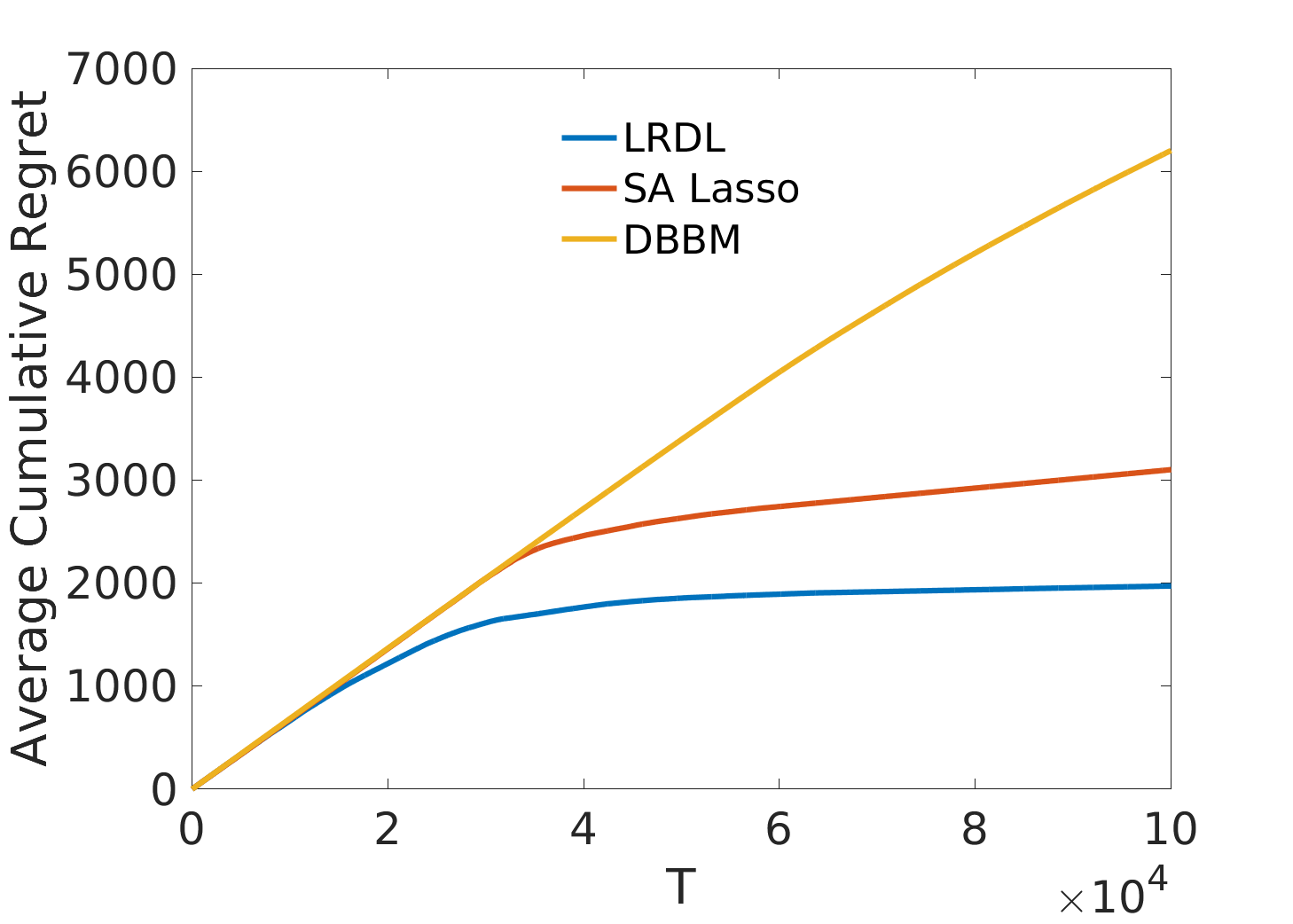}}
\subfloat[$\kappa = 0$]{\includegraphics[width=0.45\textwidth]{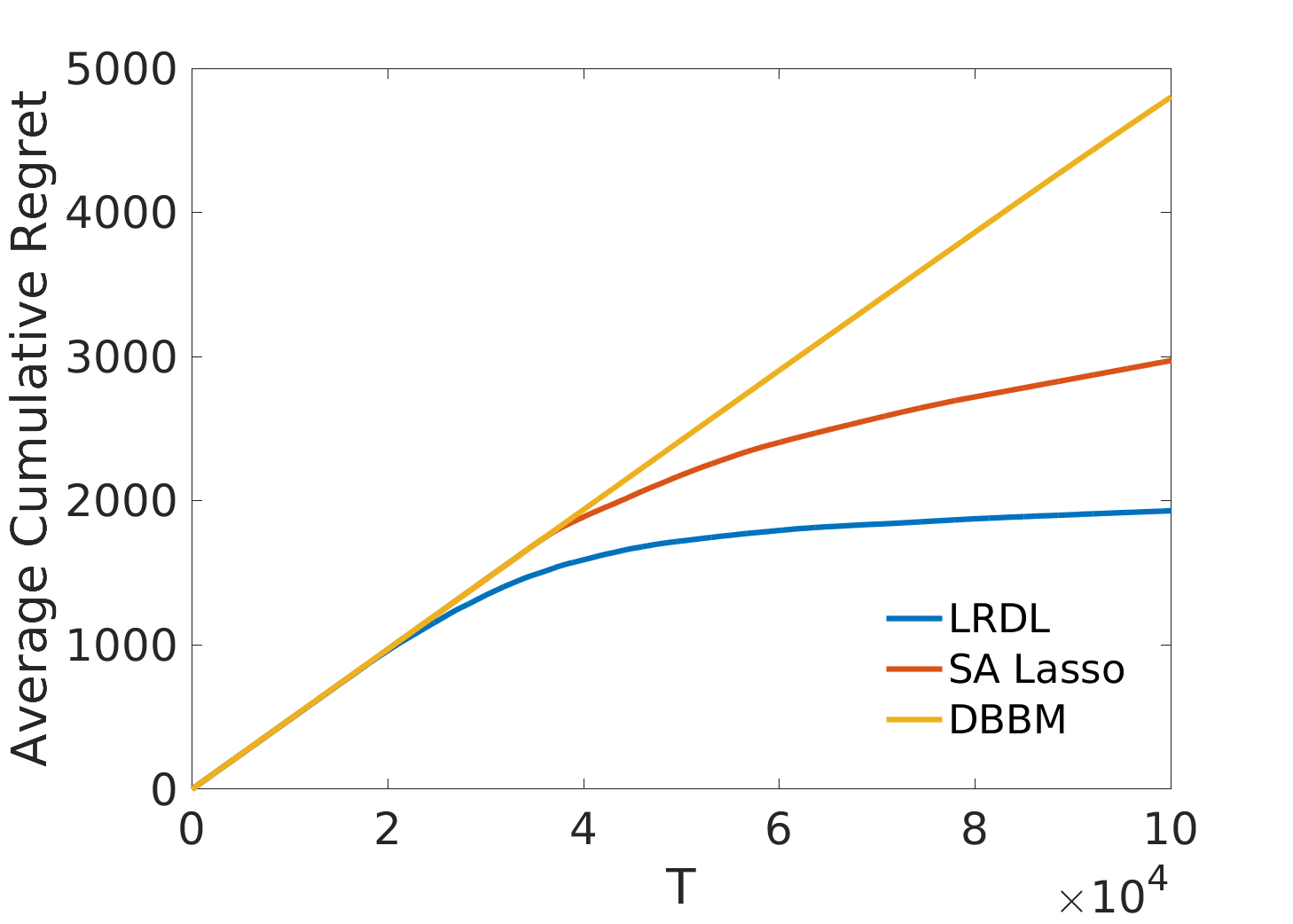}}
 \linebreak 
\caption{Cumulative regret of LRDL (Algorithm \ref{alg_LRDL}), DBBM \citep{geng2020comparison}, and SA Lasso \citep{oh2020sparsity} for high dimensional synthetic problems ($K=10, R=100$). We consider the setting with strong CA and DA effect ($\kappa=1$) and with no CA and DA effect ($\kappa=0$). The displayed statistics are averaged over 40 independently generated test problems. We observe that DBBM has superior regret performance over the benchmakrs. In particular, DBBM suffers linear regret due to lack of cross-learning.  }
    \label{fig:large_regret}
\end{figure}

\noindent\textbf{Results.} \quad As shown in Figure \ref{fig:large_regret}, the LRDL method has a superior regret performance over both DBBM and SA Lasso. For DBBM, we observe that it barely makes any progress over the 100,000 trials, and it suffers linear regret over the entire experiment. This is because DBBM learns CTRs for different CA-TA combinations independently -- with only hundreds of clicks over the entire experiment, DBBM cannot generate meaningful estimates for any of its 1,000 CTRs.

Comparing LRDL with SA Lasso, we see that LRDL outperforms SA Lasso, especially when $\kappa = 0 $. This is because SA Lasso bases its exploration on the variability coming from the context (see the relative symmetry assumption in \cite{oh2020sparsity}). However, as the feature vector contains an initial component of 1 (fixed) corresponding to the intercept term $c_0$, and as there is less randomness from other covariates ($\alpha_k,\beta_r,\gamma_{kr}$) when $\kappa=0$, the relative symmetry condition (requiring that the distribution of the context vector be relatively symmetric around the origin) fails and SA Lasso does not explore enough. When $\kappa = 1$, there is more variability from other covariates, so the performance of SA Lasso improves. A clear illustration can be found in Table \ref{tb:high}. We observe that by $T = 10,000$, all algorithms are still at the exploration stage. Shortly afterward, LRDL picks up signals and achieves more than 84.0\% of the optimal rewards through 50,000 trials; SA Lasso is outperformed by LRDL, achieving 53.4\% when $\kappa = 0$ and 82.9\% when $\kappa = 1$. DBBM only improves 3.7\% over the entire experiment when $ \kappa=0$, and 18.9\% when $\kappa=1$. The reason for SA Lasso performing better with $\kappa=1$ was explained earlier in Subsection \ref{subsec:real}.

\begin{table}[H]
\caption{Reward comparisons of LRDL (Algorithm \ref{alg_LRDL}), DBBM \citep{geng2020comparison}, and SA Lasso \citep{oh2020sparsity} for high-dimensional synthetic problems ($K=10, R=100$). Figures in the body of table show CTR in absolute terms (clicks per trial), and then as percent of maximum possible.}
\smallskip
\label{tb:high}
\centering
\begin{tabular}{l|l|ll|ll|ll|ll} \hline
\multicolumn{2}{l|}{ } & \multicolumn{2}{l|}{T = 5k} & \multicolumn{2}{l|}{T = 10k} & \multicolumn{2}{l|}{T = 50k} & \multicolumn{2}{l}{T = 100k} \\ \hline
\multirow{3}{*}{$\kappa=0$} & LRDL     & 0.012      & 21.0\%      & 0.012       & 21.3\%      & 0.052       & 84.0\%      & 0.057       & 95.3\%       \\
                         & SA Lasso & 0.011      & 19.4\%      & 0.011       & 19.0\%      & 0.032       & 53.4\%      & 0.047       & 80.3\%       \\
                         & DBBM     & 0.011      & 19.9\%      & 0.011       & 19.8\%      & 0.011       & 19.9\%      & 0.014       & 23.7\%       \\ \hline
\multirow{3}{*}{$\kappa=1$} & LRDL     & 0.018      & 20.4\%      & 0.017       & 19.7\%      & 0.075       & 89.7\%      & 0.079       & 97.3\%       \\
                         & SA Lasso & 0.014      & 19.5\%      & 0.011       & 13.8\%      & 0.068       & 82.9\%      & 0.072       & 89.9\%       \\
                         & DBBM     & 0.012      & 16.2\%      & 0.013       & 16.9\%      & 0.015       & 18.9\%      & 0.033       & 35.1\%    \\ \hline  
\end{tabular}
\end{table}

	\section{The platform's two-stage contextual bandit problem} \label{sec:twostage}

The model formulated in Section \ref{sec:problem} involves an experiment implemented by an advertiser. In this section, we introduce a new variant of the problem, which is motivated by the increased propensity of tech platforms to offer automation algorithms that optimize campaigns on behalf of their advertisers. In this variant, a platform uses an algorithm to conduct an experiment, exploiting detailed information about users that is visible to the platform but not to the advertiser. The difference between the two models thus quantifies the benefits to an advertiser from having an online platform conducting experiments on its behalf.  While any additional information about the impression can be considered, we are going to work with a specific implementation of the side information that is utilized by \cite{geng2020comparison}. In that approach, given a set of TAs, the platform constructs a \textit{minimal} partition of the user population into \textit{disjoint audience segments} (DAs) such that each of the TAs for which advertisers can purchase impressions is the union of finitely many nonoverlapping DAs. For example, if the TAs are ``San Francisco users'' and ``Male users,'' we create three DAs, ``San Francisco users, Male,'' ``San Francisco users, Not Male,” and “Non-San Francisco users, Male.'' In general, when we have $K$ TAs, the number of DAs can be as large as $(2^K-1)$.

Given that added structure, we consider the following two-stage problem. First, as in our original formulation, the algorithm chooses a TA from the available set. Second, a user is randomly selected from that TA, \textit{the user’s DA is observed}, and then\textit{ a CA is selected for display based on the observed DA}. In the physical implementation of this sequence, it is the platform that must actually select the CA based on the observed DA, but we suppose that it does so by following the contingent instructions provided by the advertiser (see below for elaboration).

Of course, this two-stage formulation assumes that the user's DA is indeed observable by the platform, and that the choice of a CA can be based on that observation. Using language that is standard in the bandit literature, we then have a \textit{contextual} bandit model, where the user DA observed in a given trial constitutes the context in which that trial's CA is selected.

The subsections below describe precisely our model of the two-stage problem, describe a modified version of our LRDL algorithm for its solution, and compare the outcomes realized by the advertiser with those realized in our original single-stage formulation.

\subsection{LRDL framework for the two-stage problem} \label{subsec:2stagealg}
   
With additional information, the platform constructs a regression model at the DA level. Specifically, assume the click probability of a user from DA $j$ after seeing CA $r$ is $p_{jr}$. Again we consider a logistic regression model, but now with $p_{jr} =\left(1+\exp(-c_0-\alpha_j -\beta_r -\gamma_{jr})\right)^{-1} =  \left(1+\exp(-\theta^\top \phi(j,r))\right)^{-1}$, where 
\eqss{ 
\phi(j,r) &= \big( 1, \one_{1}(j),\one_{2}(j), \cdots, \one_{J}(j), \one_{1}(r), \one_{2}(r), \cdots, \one_{R}(r), \one_{1,1}(j,r), \cdots, \one_{J,R}(j,r) \big)^\top, \text{and}\\
\theta &= \big(c_0, \alpha_1, \cdots,  \alpha_J,  \beta_1, \cdots,\beta_R, \gamma_{11},    \cdots, \gamma_{JR} \big).
} 
Similar to Section \ref{sec:method}, we impose zero sum constraints $ \sum_j\alpha_j= \sum_r \beta_r=0$ and $\sum_r \gamma_{jr} = \sum_{j} \gamma_{jr}$ for all $j$ and $r$. With this convention, the feature vector has $d=JR$ components, where $J$ can be as large as $(2^K-1)$ (see Subsection \ref{subsec:2stagenumerical}). Thus, we may see a dramatic increase in problem size due to the two-stage structure. For example, with $K=5$ TAs  and $R=10$ CAs, a single-stage model has $d=50$ parameters to estimate, while the corresponding two-stage model may have $d\approx 300.$

At the beginning of each trial $t,$ the algorithm specifies the TA $k_t$ to purchase, as well as a DA-level policy $f_t: \text{DA} \rightarrow \text{CA}$. The policy is calculated by first sampling a vector $ \tilde{\theta}$ from a ``posterior'' probability distribution, and then computing $\tilde{p}_{jr} = \left(1+\exp(-\tilde{\theta}^\top \phi(j,r)) \right)^{-1}$. We then set
\eqss{
f_{t}(j) &: = \argmax_r \tilde{p}_{jr}\text{~for every~} j, \text{and}\\
k_t &: = \argmax_{k} \sum_j \PP(j|k)~ \tilde{p}_{j f_{t}(j)},}
where $\PP(j|k)$ is the conditional probability that a user drawn at random from TA $k$ belongs to DA $j,$ which is treated here as a known constant. The elements of the proposed two-stage LRDL method are specified in Algorithm \ref{alg_dblasso3}. For simplicity, we denote $\phi(j, r) $ by $x_{jr}.$ Table \ref{tb:2stageDifference} summarizes the differences between the advertiser's single-stage problem and the platform's two-stage problem. 

\begin{table}[H]   
    \caption{Problem formulations for the advertiser's single-stage problem and the platform's two-stage problem.}
    \smallskip
    \label{tb:2stageDifference}
        \centering
         \begin{adjustwidth}{-1cm}{}
        \begin{tabular}{|c| c|c| c |c |c |}
         \hline
           &  decision maker & Information& Effective dim & Control & Objective  \\ 
         \hline
         Single-stage model & Advertiser & $y_t$ &   $K\cdot R$  & $(k_t, r_t)$   & $\max \sum_{t=1}^T  p_{k_t r_t} $ \\
         Two-stage model & Platform &  $y_t, j_t$ &  $J\cdot R \sim O(2^K \cdot R)$ & $k_t, f_t: DA \rightarrow CA $ & $  \max \sum_{t=1}^T \sum_j \mathbb{P}(j|k_t)~ p_{j f_t(j)} $  \\
     \hline 
        \end{tabular}     
    \end{adjustwidth}

\end{table}

\begin{algorithm}[H]
 \label{alg_dblasso3}
 \caption{LRDL Algorithm for the two-stage problem}
\KwIn{ $ c, \hat{\theta}_d = [0, 0, \cdots, 0] \in \RR^d, \hat{\Sigma}_d = \vI_d$\\
Set $ d= JR, \lambda = c \sqrt{\log d/t}$.
}
\For{$t=1,2, \cdots, T$}{
Sample $ \Tilde{\theta} \sim \cN(  \hat{\theta}_d  , \rho^2\hat{\Sigma}_d)$ \\
Submit $ f_{t}(j) : = \argmax_r  \left(1+\exp(- \Tilde{\theta} ^\top x_{jr}) \right)^{-1}, \forall j,$  and\\
$k_t : = \argmax_{k} \sum_j \PP(j|k)~   \left(1+\exp(- \Tilde{\theta} ^\top x_{jr}) \right)^{-1}$ to the platform.\\
Platform sample one user $j_t$ from TA $k_t$, display CA $f_t(j_t)$, and observe $y_t$.\\
Update $ \cH_t \leftarrow \cH_{t-1} \cup \{(x_t, y_t) \}, $ where $x_t=\phi(j_t, r_t)$.\\ 

Compute  $\hat{\theta}_l  = \argmin_{\theta} \frac{1}{t} \sum_{\tau=1}^t [ \log \left(1+\exp(x_\tau^\top\theta)\right) - y_{\tau} x_{\tau}^\top \theta  ] + \lambda ||\theta||_1$, and
$ \hat{\Sigma} = \frac{1}{t} \sum_{\tau=1}^t  \left( 1+\exp(x_\tau^\top\hat{\theta}_l) \right)^{-1} \left( 1+\exp(-x_\tau^\top\hat{\theta}_l) \right)^{-1} \cdot x_{\tau} x_{\tau}^\top$.

Update $\hat{\theta}_d  \leftarrow  \hat{\theta}_l + \frac{1}{t} \hat{\Sigma}^{-1} \sum_{\tau =1}^t \left(y_\tau -  \left( 1+\exp(-x_\tau^\top\hat{\theta}_l) \right)^{-1} \right) x_\tau,$ and $ \hat{\Sigma}_d \leftarrow \hat{\Sigma}^{-1} / t.$

}

\end{algorithm}

\subsection{Numerical experiments} \label{subsec:2stagenumerical}

The numerical experiments described in this subsection show that our LRDL method remains effective in the more complex two-stage environment. 


\subsubsection{Real data}

In our test problem based on real data (see Subsection \ref{subsec:real}), we have 4 TAs, corresponding to 11 DAs, and 26 CAs. That is, the single-stage model has problem size $d=4\times26 =104$, whereas the two-stage problem has $d=11\times26=286$. In appendix \ref{app:real} we display the click-through rates $p_{jr}$ for all DA-CA pairs $(j,r)$ in Table \ref{tb:2stage_CTR}, and display the conditional probabilities $\PP(j|k)$ for all DA-CA pairs $(j, k)$ in Table \ref{tb:TADA}. The data in appendix \ref{app:real} is perturbed for confidentiality.

Table \ref{tb:2stageReal} shows the CTR performance of our LRDL method in both the single-stage and two-stage problems, and also that of the DBBM method used by \cite{geng2020comparison}, which has an obvious extension to the more complicated two-stage setting.

\begin{table}[H]
\caption{Reward comparison of the two-stage problem and single-stage problem on real data. Figures in the body of the table show CTR in absolute terms (click per trial), and then as percent of the maximum possible.}
\smallskip
\label{tb:2stageReal}\centering
\begin{tabular}{l|ll|ll|ll|ll}  \hline
                  & \multicolumn{2}{l|}{T=5k} & \multicolumn{2}{l|}{T=10k} & \multicolumn{2}{l|}{T=50k} & \multicolumn{2}{l}{T=100k} \\ \hline
Single-stage LRDL & 0.003      & 18.6\%      & 0.006       & 37.4\%      & 0.015       & 95.8\%      & 0.015       & 96.0\%       \\
Single-stage DBBM & 0.002      & 13.2\%      & 0.002       & 13.9\%      & 0.009       & 53.9\%      & 0.014       & 85.9\%       \\
Two-stage LRDL    & 0.005      & 21.3\%      & 0.010       & 41.2\%      & 0.019       & 80.3\%      & 0.022       & 93.2\%       \\
Two-stage DBBM    & 0.004      & 16.0\%      & 0.005       & 21.8\%      & 0.015       & 64.6\%      & 0.020       & 84.1\% \\ \hline     
\end{tabular}
\end{table}

\subsubsection{Synthetic data} 

We consider 10 CAs and 5 TAs, which implies 31 ($2^5-1$) DAs. For each DA, we randomly draw a weight $w_j \sim \text{Unif}[0,1]$, then use those weights to calculate the conditional probabilities $\PP(j|k)$ as follows:
\eqs{
 \PP(j|k) = \frac{w_j}{\sum_{j \in \text{TA}k} w_j}~\text{~for all~} j\in \text{TA}~k.}
 Suppose, for example, that $K=2$ and assume DA1$\subset$TA1, DA2 $\subset$ TA1$\cap$TA2, DA3$\subset$TA2. If $w_1 = 0.1, w_2 = 0.2, w_3 = 0.3$, we have $\PP(DA1|TA1) = \frac{1}{3}, \PP(DA2|TA1)=\frac{2}{3},  \PP(DA2|TA2)= \frac{2}{5} $ and $  \PP(DA3|TA2) = \frac{3}{5}$. 
 
 This approach helps differenciate different TAs. As in Section \ref{subsec:synthetic}, the true CTR value of each CA-DA combination is generated using the following obvious variant of formula \eqref{eq:pkr}:
\eqs{ \label{eq:pjr}  p_{jr} = (1+\exp(-c_0-\kappa\alpha_j-  \kappa\beta_{r} - \gamma_{jr}))^{-1}.} 
The parameters $c_0, \alpha_j,\beta_r$ and $\gamma_{jr}$ are referred to as the \textit{constant term}, a \textit{DA effect}, a \textit{CA effect}, and an \textit{interaction effect}, respectively. Again, we set $c_0=5$ and all parameters ($\alpha_j, \beta_r, \gamma_{jr}$) are set to 0 except a randomly selected subset. We randomly select five $\alpha_j$ components , five $\beta_r$ components, and twenty $\gamma_{jr}$ components to be nonzero, the values of which are drawn independently from a uniform distribution on $[-1,1].$ The multiplier $\kappa$ determines the strength of both DA and CA effects. When $\kappa$ is high, it is likely that the same CA will be the best for all DAs, so having extra DA information may not be helpful.

\noindent\textbf{Algorithm inputs.} \quad For LRDL, we set $c = 2$ and $\rho= 0.1.$

\noindent\textbf{Results.} \quad Table \ref{tb:2stageCompare} summarizes the comparison results when $K = 5, J = 31, R = 10$. Further discussion is deferred to Subsection \ref{sec:val_2stage}.




\begin{table}[H]
\caption{Reward comparison of the two-stage problem and single-stage problem on synthetic data. Figures in the body of the table show CTR in absolute terms (click per trial), and then as percent of the maximum possible.}
\smallskip
\label{tb:2stageCompare}\centering
\begin{tabular}{l|l|ll|ll|ll|ll}  \hline
                         &                   & \multicolumn{2}{l|}{T=5k} & \multicolumn{2}{l|}{T=10k} & \multicolumn{2}{l|}{T=50k} & \multicolumn{2}{l}{T=100k} \\ \hline
\multirow{4}{*}{$\kappa=0$} & Single-stage LRDL & 0.008      & 70.3\%      & 0.008       & 72.2\%      & 0.010       & 89.2\%      & 0.010       & 93.1\%       \\
                         & Single-stage DBBM & 0.007      & 68.0\%      & 0.007       & 68.3\%      & 0.008       & 71.9\%      & 0.008       & 76.8\%       \\
                         & Two-stage LRDL    & 0.008      & 59.0\%      & 0.010       & 66.0\%      & 0.013       & 89.8\%      & 0.014       & 94.1\%       \\
                         & Two-stage DBBM    & 0.008      & 54.0\%      & 0.008       & 56.0\%      & 0.010       & 72.0\%      & 0.012       & 79.2\%       \\ \hline
\multirow{4}{*}{$\kappa=1$} & Single-stage LRDL & 0.013      & 68.8\%      & 0.016       & 83.3\%      & 0.018       & 93.0\%      & 0.018       & 93.4\%       \\
                         & Single-stage DBBM & 0.010      & 52.4\%      & 0.011       & 56.2\%      & 0.015       & 77.4\%      & 0.017       & 86.5\%       \\
                         & Two-stage LRDL    & 0.011      & 48.4\%      & 0.013       & 56.1\%      & 0.019       & 83.8\%      & 0.021       & 91.5\%       \\
                         & Two-stage DBBM    & 0.010      & 42.2\%      & 0.011       & 45.8\%      & 0.015       & 64.1\%      & 0.017       & 73.4\% \\ \hline      
\end{tabular}
\end{table}

\subsection{Performance gains from the platform's conduct of the experiment} \label{sec:val_2stage}

To repeat, the basic model introduced in Section \ref{sec:problem} describes the common problem faced by an advertiser who conducts experiments on its own. On the other hand, the two-stage model (Section \ref{sec:twostage}) is suitable for the case where an advertiser can have an online platform making TA and CA choices on its behalf, using programmed logic provided by the advertiser and fine-scale user information available only to the platform. The difference between the two models thus quantifies the benefits that may accrue to an advertiser from having an online platform conducting experiments on its behalf. 

\begin{figure}[H]
    \centering
    \includegraphics[width=0.45\textwidth]{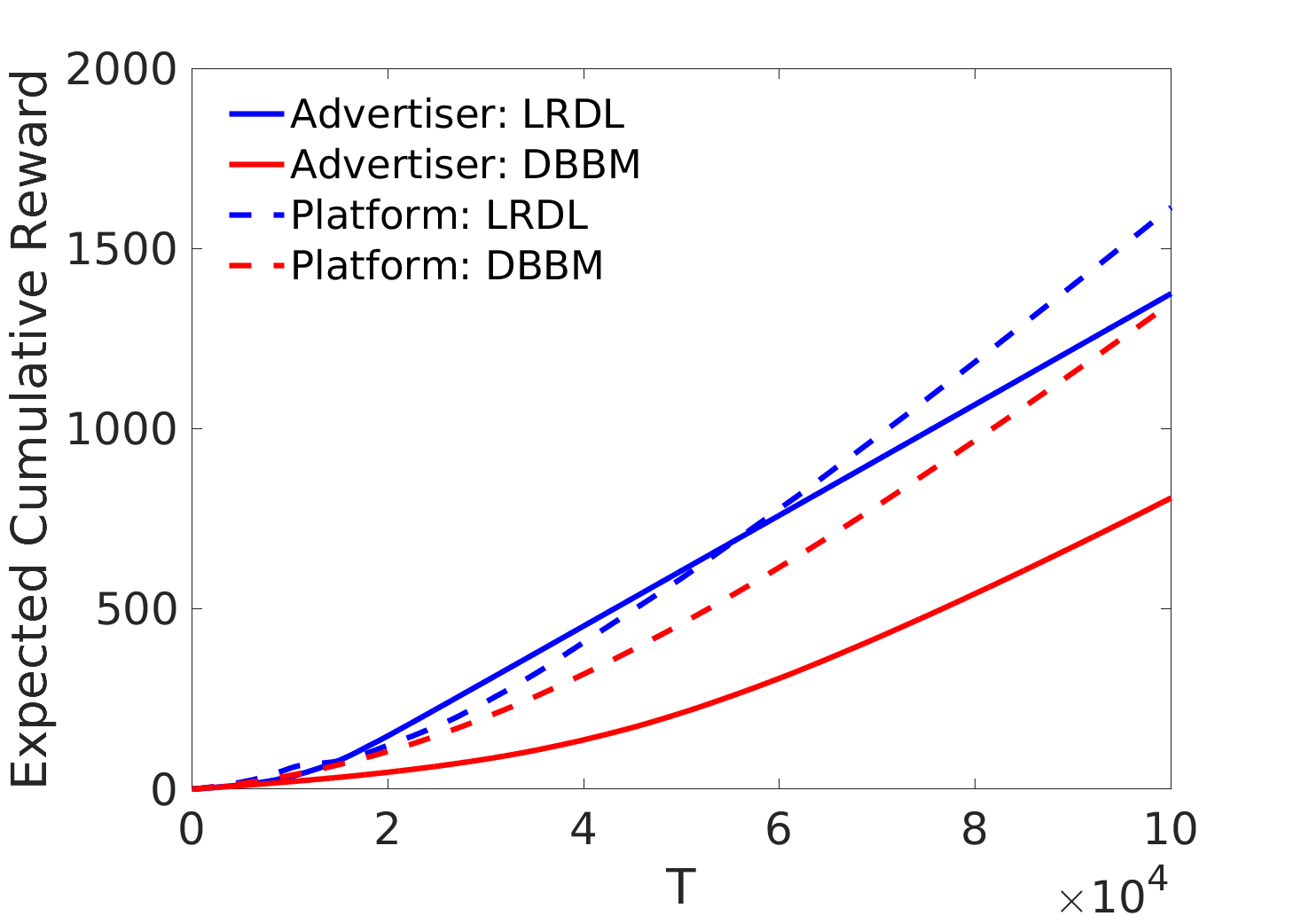}
 \linebreak 
\caption{Cumulative reward of LRDL (Algorithm \ref{alg_LRDL}) and DBBM \citep{geng2020comparison} on real data under two learning regimes (advertiser's problem vs. platforms problem). Figure shows that, for both algorithms, it is beneficial to have the platform conduct experiments on advertiser's behalf.}
    \label{fig:2scenario_real}
\end{figure}

For our test problem based on real data, Figure \ref{fig:2scenario_real} shows the expected cumulative reward achieved by LRDL and DBBM in those two learning regimes. Here ``expected cumulative reward'' is defined to mean
\eqs{\label{eq:reward}
\EE\left( \sum_{t=1}^T p_{k_t r_t}\right)~\text{for the advertiser's problem,}
}
and to mean
\eqs{
\label{eq:platform_reward}
\EE\left( \sum_{t=1}^T  \sum_{j} \PP(j|k_t) p_{j_t r_t}\right)~\text{for the platform's problem.}
}
As in the implementation of \eqref{eq:regret}, these expectations are computed via Monte Carlo replications that use the click probabilities $p_{kr}$ or $p_{jr}$ as input data. For both algorithms, the platform achieves substantially higher rewards than the advertiser. Two factors to consider when evaluating these results are the following. First, extending from the single-stage to the two-stage setting does not greatly increase the dimension of our test problem (104 versus 286), and that is helpful for the platform's performance. Second, examination of the true parameter values for our test problem shows that the maximum possible reward rate in the platform's problem is 0.023 per trial, compared to 0.016 per trial in the advertiser's problem, which is a 44\% increase.

Results reported earlier for our synthetic test problems further highlight differences between the two learning regimes. Table \ref{tb:2stageCompare} shows that the value added by the platform (in the form of per period reward) is problem-dependent and algorithm dependent. When using the LRDL algorithm, independent of $\kappa,$ the rewards in the platform's problem are higher than in the advertiser's. When using the DBBM algorithm, the reward comparison varies depending on the potential for reward improvement controlled by $\kappa$. If $\kappa =0,$ then the platform's setting is more favorable than the advertiser's. However, if $\kappa = 1,$ the platform's setting is distinctly \textit{less} favorable than the advertiser's. That is, providing ex-post information on user DA, and allowing for a dynamic CA response to that information, actually \textit{decreases} the advertiser’s expected revenue over 100,000 trials.

This behavior is driven by whether the algorithm enforces cross-learning and feature selection (the process of selecting a subset of features for use in the model and setting the values of other variables to zero), and by whether the added value (in theory) is significant. Recall that $p_{jr} = (1+\exp(-c - \kappa \alpha_j - \kappa\beta_r - \gamma_{jr}))^{-1}$. When $\kappa = 1$, there are strong CA and DA effects, and it is likely that the same CA will be best for all DAs, so there isn't much additional value in dynamically selecting CA contingent on the observed DA. In contrast, when $ \kappa =0$ we have $p_{jr} = ({1+\exp(-c - \gamma_{jr})})^{-1}$, each DA may map to a different best CA, and the two-stage problem has a much higher optimal reward. Though the theoretical optimal value is higher in the two-stage problem, the problem is also of a much higher dimension (310 CTRs vs. 50 CTRs), where cross-learning and feature selection becomes crucial for an algorithm to exploit the added value. Indeed, when the potential is large ($\kappa = 0$), both DBBM and LRDL are able to achieve a higher reward. However, when the potential is small ($\kappa =1$), DBBM is led astray by the excessive information and fails to pick up important signals, because it assumes independent arms. LRDL correctly handles the high-dimensional problem by sharing information between arms and by feature selection, thus obtaining a higher reward in the two-stage setting despite the small potential to start with. 

Of course, one naturally expects that more information and more decision-making flexibility will be beneficial. However, our simulations show that the benefit realized depends on the sophistication of the logic used to process the additional information and respond to it. Without cross-learing, a naive algorithm may be overwhelmed by the additional information, hence overexplores suboptimal options and leads to worse performance. 
	
\section{Discussion} \label{sec:discussion}

The LRDL method that we have proposed incorporates several types of refinement, and in this section we show via simulation how those refinements improve the algorithm's performance. Subsection \ref{subsec:val_refinement} shows the importance of including judiciously selected interaction terms in our logistic regression model; Subsection \ref{subsec:val_debiasing} explores the value added by debiasing our lasso parameter estimates; and Subsection \ref{subsec:adjustable_exp} studies the value added by adjustable exploration.

\subsection{Performance gains from inclusion of interaction terms }\label{subsec:val_refinement} 

Inclusion of the interaction terms $\gamma_{kr}$ in our regression model \eqref{eq:logistic_model} greatly increases the number of model parameters to be estimated, and therefore the complexity of our analysis. Given the limited time horizon and scarce feedback that are typical in online advertising, one may wonder if it is worthwile to adopt such a complicated model structure. Instead, a simple and natural probability model we may adopt is the two-way additive model mentioned in Subsection \ref{subsec:pos_features}, namely, \eqs{\label{eq:2way}
p_{kr} =\left(1+ \exp(-c_0 - \alpha_k - \beta_r)\right)^{-1}.}
A drawback of this simplified model is its potential bias, but it still may give better performance over a limited time horizon. We therefore compare LRDL performance with that of an algorithm based on \eqref{eq:2way}, and for that purpose we choose the GLM-UCB algorithm of \cite{li2017provably}. We choose this standard of comparison because (a) with the interaction terms excluded in our logistic regression model, there are only $K+R-1$ regression parameters to be estimated, so the problem is low-dimensional, and (b) of the existing methods considered in Section \ref{subsubsec:low}, GLM-UCB gave by far the best performance on low-dimensional problems  (see Figure \ref{fig:small_regret}).

We implement the comparison first in our test problem based on real data (see Subsection \ref{subsec:real}), and then on a family of synthetically generated test problems as in Subsection \ref{subsec:synthetic}, with $ K = 10$, $R = 50$ and $\kappa \in \{0, 1\}$. For LRDL, the feature vector then has dimension $K\times R= 500$, and for GLM-UCB, it has dimension $K+R-1=59$. For LRDL, we set $c= 12 $ and $\rho=0.1$; for GLM-UCB, we set $\alpha=1$. 

\begin{figure}[]
    \centering
     \subfloat[Real data]{\includegraphics[width=0.45\textwidth]{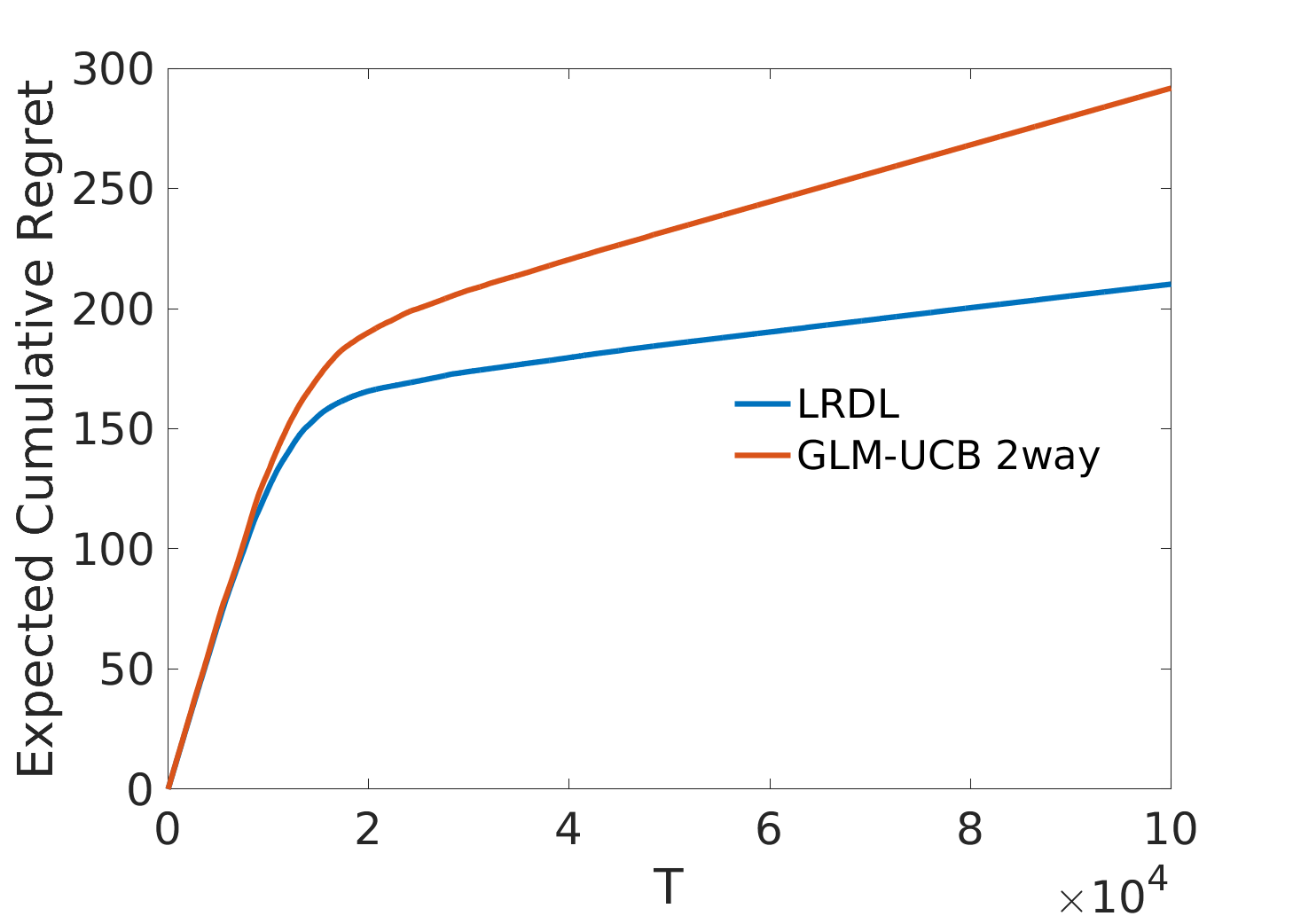}}\\
     \subfloat[Synthetic data, $\kappa=0$]{\includegraphics[width=0.45\textwidth]{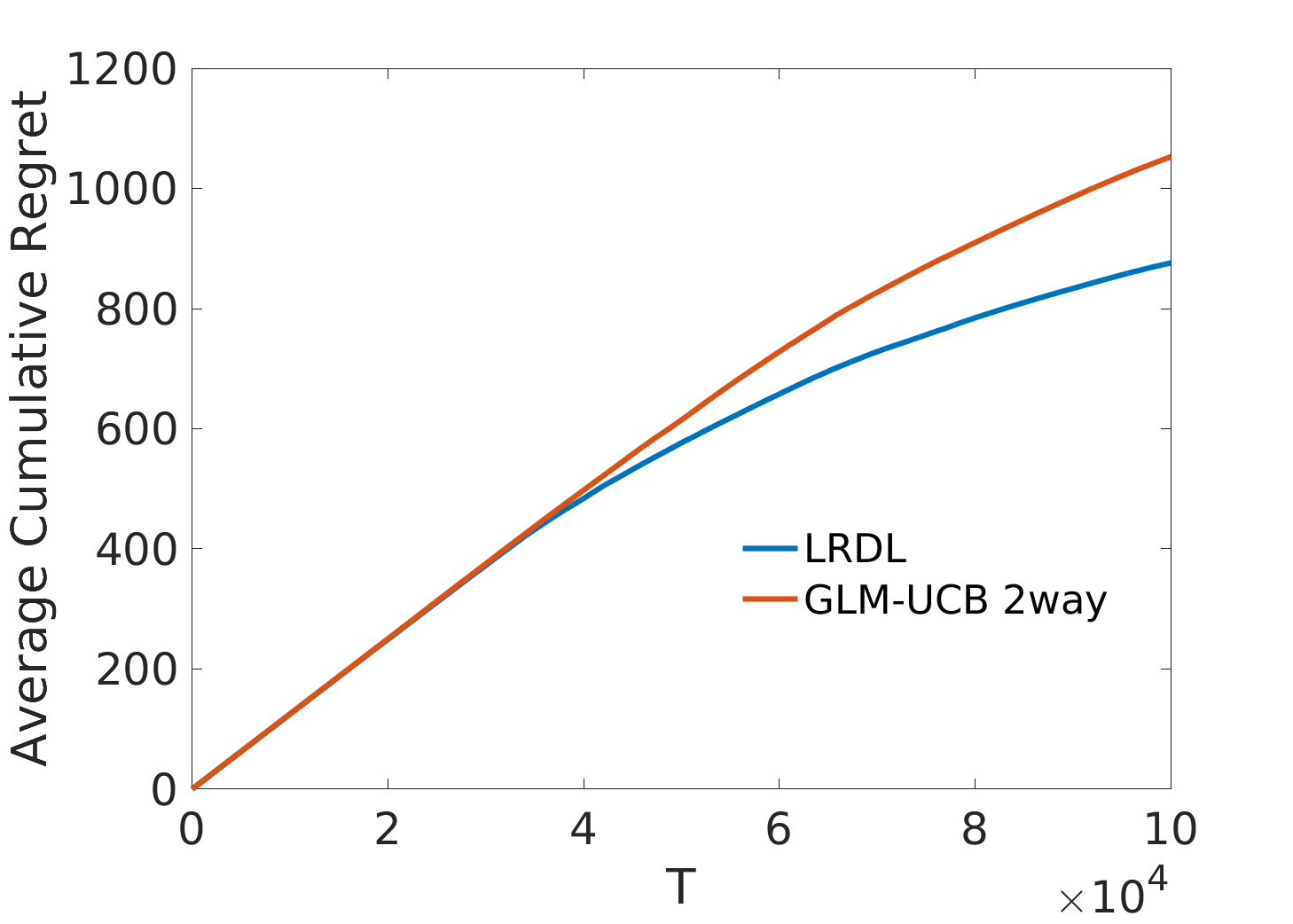}}
    \subfloat[Synthetic data, $\kappa=1$]{\includegraphics[width=0.45\textwidth]{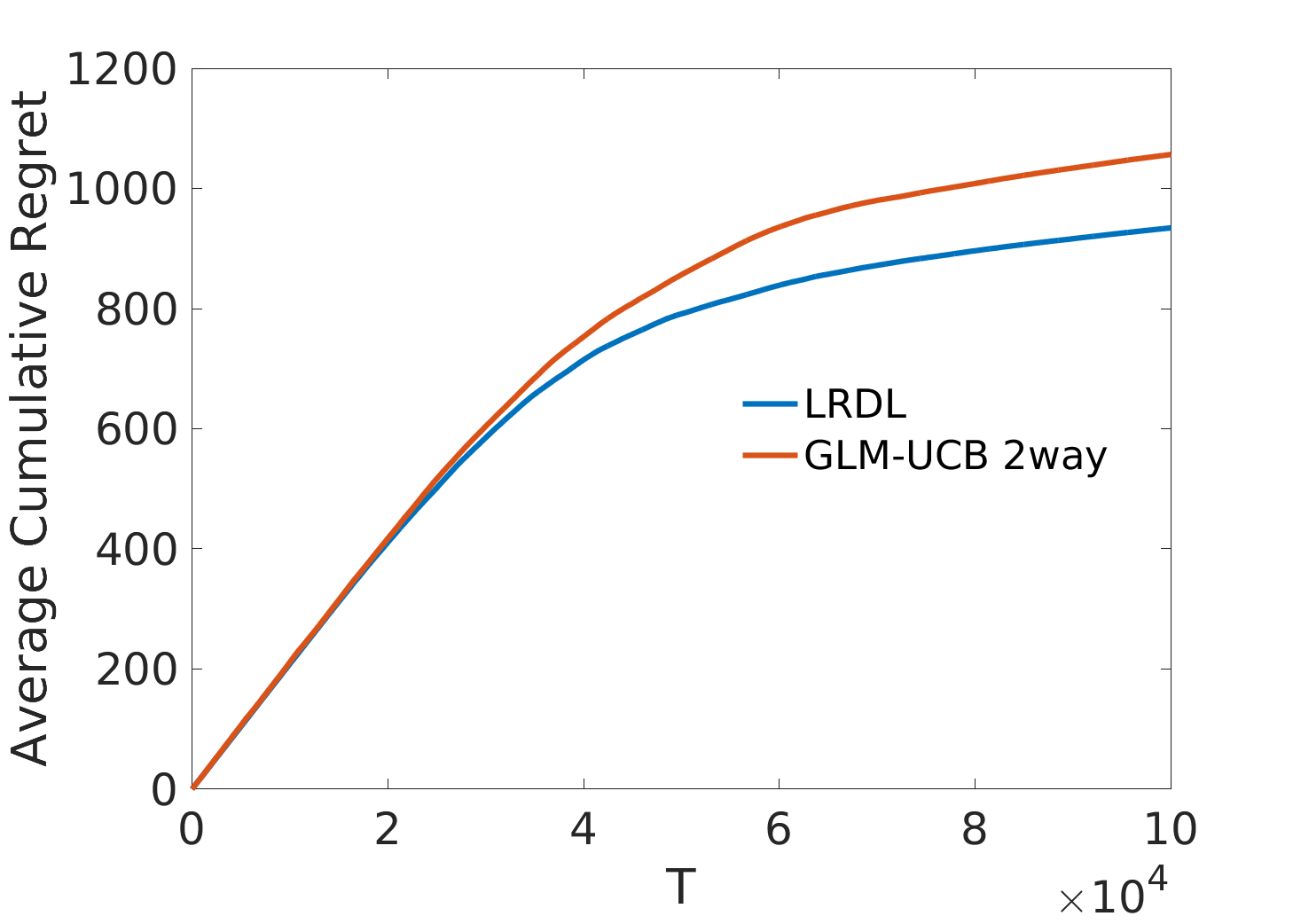}}
~\linebreak
\caption{Comparative performance using models with and without interactions terms, based on cumulative regret in our test problem using real data ($K=4, R=26$) and synthetically generated test problems ($K = 10$ and $R=50 $). For synthetic data, we consider the setting with strong CA and TA effects ($\kappa=1$) and with no CA and TA effects ($\kappa=0$). Figure compares the GLM-UCB (2-way additive model) and LRDL (2-way additive model with interaction terms). The displayed statistics are averaged over 40 independently generated trials.}
\label{fig:val_refinement}
\end{figure}

Figure \ref{fig:val_refinement} shows cumulative regret achieved by GLM-UCB and by LRDL. We observe postive gains from including interaction terms in our model and then using algorithmic feature selection (that is, using $\ell_1$ regularization to find a subset of variables to use for the model, while setting the values of other variables to zero). Specifically, when implemeted on real data (Figure \ref{fig:val_refinement}(a)), the average cumulative regret using LRDL is 210, a 28\% reduction from the average cumulative regret of 292 using GLM-UCB. When implemented on synthetic data, the performance gain by using LRDL is higher with $\kappa=0$ (Figure \ref{fig:val_refinement}(b)) than with $\kappa=1$ (Figure \ref{fig:val_refinement}(c)). An important factor contributing to this difference is that the two-way additive model \eqref{eq:2way} has a substantial built-in bias when $\kappa=0$, causing cumulative regret to increase almost linearly as $T$ increases. When $\kappa=1$, the two-way additive model has less built-in bias (compared to $\kappa=0$): the average cumulative regret using GLM-UCB is reduced slightly to 1057, but the cumulative regret using LRDL is still 14\% less at 911.

\subsection{Essential role of debiasing}\label{subsec:val_debiasing}

We now compare our LRDL method with what will be called  Logistic Regression with Biased Lasso (LRBL). The latter algorithm is identical to LRDL except for the following: the parameter distribution used for posterior sampling at any given stage is a Gaussian distribution with mean vector equal to the (potentially biased) lasso parameter estimate, and with the same covariance matrix as in LRDL. The LRBL method is summarized in Algorithm \ref{alg:LRBL}.

\begin{algorithm}[H]
\KwIn{ $ c, \rho, \hat{\theta}_d  = [0, 0, \cdots, 0] \in \RR^d, \hat{\Sigma}_d = \vI_d$\\
Set  $d = KR, \lambda = c \sqrt{\log d/ t}$.
}

\For{$t=1,2, \cdots, T$}{
Sample $ \Tilde{\theta} \sim \cN(  \theta_l , \rho \Sigma_d)$ \\
Select $ (k_t, r_t) = \argmax_{k, r}  \left(1+ \exp\left(-\Tilde{\theta}^\top \phi(k,r) \right)\right)^{-1} $.\\
Platform sample one user from TA $k_t$ and display CA $r_t$, observe $y_t$.\\
Update $ \cH_t = \cH_{t-1}\cup \{(x_t, y_t)\}$, where $x_t =\phi(k_t,r_t)$.\\  
Update $\hat{\theta}_l  \leftarrow \argmin_{\theta} \frac{1}{t} \sum_{\tau=1}^t [ \log \left(1+\exp(x_\tau^\top\theta)\right) - y_{\tau} x_{\tau}^\top \theta  ] + \lambda ||\theta||_1$\\
Update  $ \hat{\Sigma}_d \leftarrow \left(\sum_{\tau=1}^t  \left( 1+\exp(x_\tau^\top\hat{\theta}_l) \right)^{-1} \left( 1+\exp(-x_\tau^\top\hat{\theta}_l) \right)^{-1} \cdot x_{\tau} x_{\tau}^\top\right)^{-1}$.
}
\caption{LRBL Algorithm}
 \label{alg:LRBL}
\end{algorithm}

\begin{figure}[H]
    \centering
\includegraphics[width=0.45\textwidth]{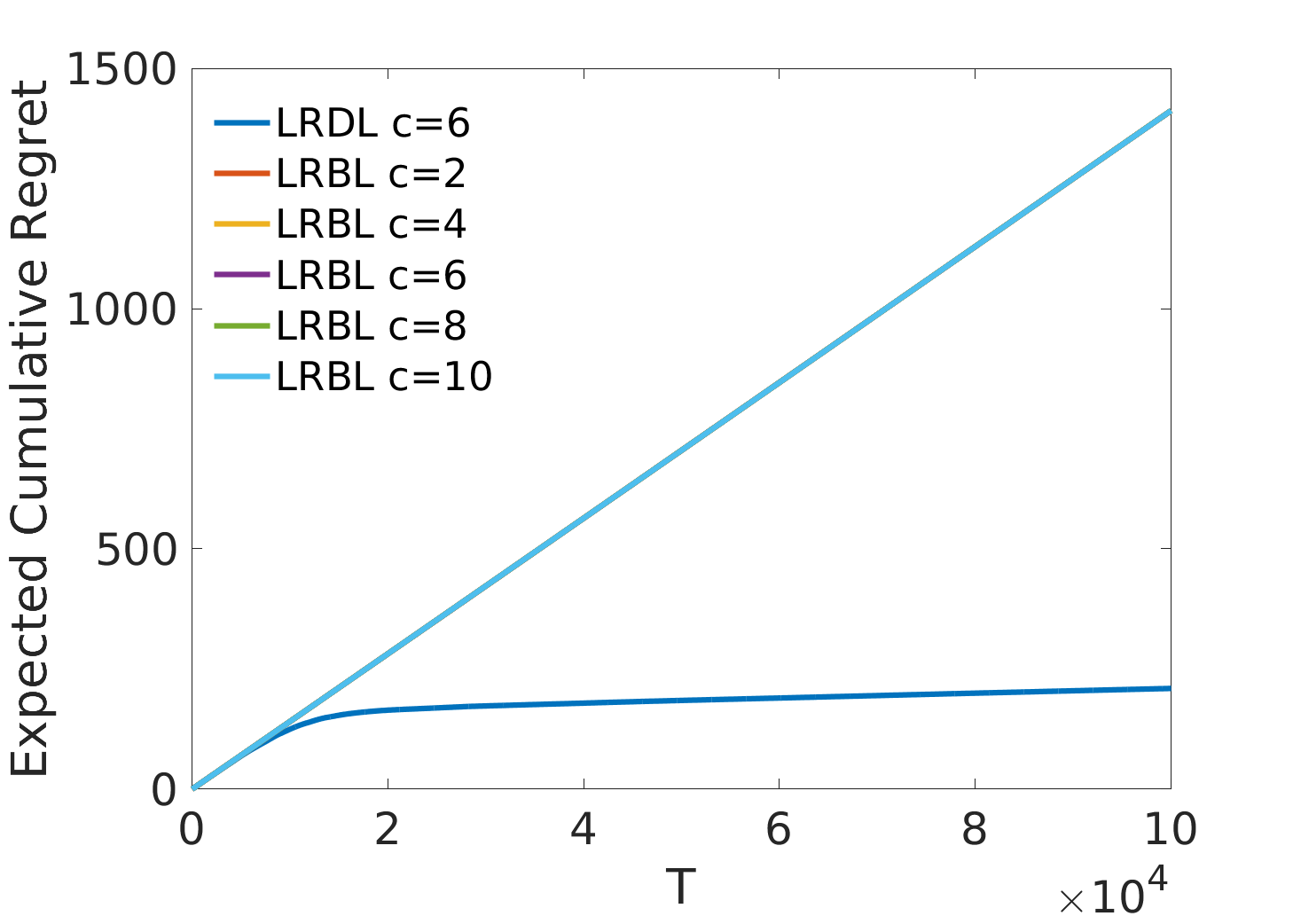}~
\linebreak
    \caption{Performance comparison between LRDL and LRBL: Expected cumulative regret in our test problem based on real data ($K=4, R= 26$). The displayed statistics are averaged over 40\ replications.}
    \label{fig:debiasing}
\end{figure}

Figure \ref{fig:debiasing} compares the expected cumulative regret of LRDL and LRBL, with different tuning parameters, in our test problem based on real data. We see that LRDL clearly outperforms LRBL. Moreover, without debiasing, LRBL produces linear regret over the entire experiment. Although its use of the covariance matrix $\hat{\Sigma}_d$ is dubious, that is clearly not the core problem with the LRBL algorithm. Rather, LRBL fails to approach the true optimum, and in fact fails to make significant progress over time, simply because it bases its reward estimates on biased parameter estimates.

\subsection{Performance gains using adjustable exploration} \label{subsec:adjustable_exp}

\begin{figure}
    \centering
    \includegraphics[width=0.45\textwidth]{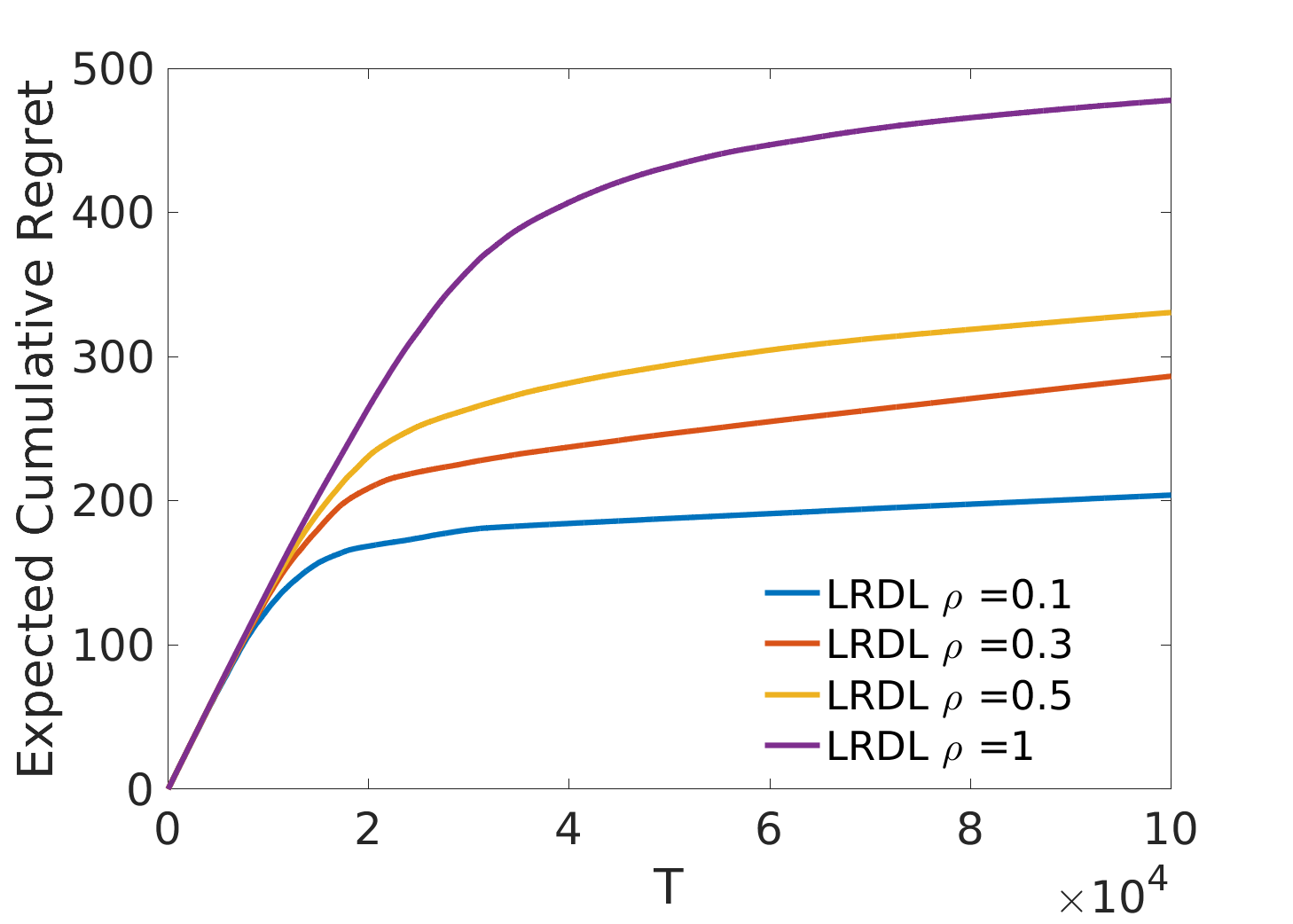}
    \caption{Expected cumulative regret for different values of the exploration index $\rho$ on our test problem based on real data. We find $\rho$ achieves the lowest regret.}
    \label{fig:explore_real}
\end{figure}

We now present a simulation study that shows how varying the adjustable exploration parameter $\rho$ affects the performance of our LRDL algorithm. Specifically, all $\rho$ values in the set $ \{0.02, 0.05, 0.07, 0.1, 0.3, 0.5, 0.7, 0.9, 1\}$ were considered. First, Figure \ref{fig:explore_real} shows the simulation result in our test problem based on real data. We find that $\rho=0.1$ achieves lowest regret, which reduces regret by about 60\% compared with standard Thompson sampling ($\rho=1$).

Next, we evaluate how the effect of $\rho$ depends on the problem dimension (number of TA-CA combinations) using synthetic test problems. Fixing $K=10$, we vary the number of creatives $R \in \{5, 10, 20, 50\}$, hence the problem dimension $d = KR \in \{50, 100, 200, 500\}$. Test problems are generated using a similar approach as in Section \ref{sec:Numerical}. For all test problems, we fix the number of non-zeros in $\theta$ and use $\kappa = 1$. For each dimension $d$, we generate 40 test problems and report the average of the cumulative regret for algorithms with different $\rho$. The CTRs of each test problem are generated via \eqref{eq:pkr}, where only a randomly chosen subset of covariates ($\alpha, \beta,\gamma$) are set to nonzero values. Specifically, we randomly select five $\alpha_k$ components, five $\beta_r$ components, and ten $\gamma_{kr}$ components to be non-zeros, and draw their values independently from a uniform distribution on [-1,1].

Table \ref{tb:expInx_regret} displays the cumulative regret of LRDL with different values of $\rho$, all for a horizon length of $T= 100,000$. Star (*) indicates the $\rho$ value that achieves the lowest regret. We observe that when $d$ is large, a smaller $\rho$ yields a lower regret. Classical Thompson sampling ($\rho=1$) is generally dominated by ``greedier'' Thompson sampling (smaller $\rho$). Adjusting the exploration index $\rho$ adds great value: as we see in Table \ref{tb:expInx_regret}, when $d=50$, cumulative regret with $\rho=1$ is almost triple that with $ \rho = 0.05$. 

\begin{table}[H]
\caption{Average cumulative regret for different values of the exploration index $\rho$, and for different problem dimensions $d$. Star (*) marks the lowest regret for each dimension. For example, $\rho=0.1$ achieves the lowest regret when $d=500$.}
\smallskip
\centering{
\begin{tabular}{l|lllllllllll}\hline
$\kappa=1$ & 0.01   & 0.03   & 0.05   & 0.07   & 0.09   & 0.1    & 0.3    & 0.5    & 0.7    & 0.9    & 1      \\ \hline
d=50    & 586.7  & 399.6  & 424.7  & 362.1  & 308.5  & 311.9  & 253.2  & 201.7*  & 231.4  & 285.3  & 302.3  \\
d=100   & 1089.0 & 663.2  & 602.1  &554.3 & 461.4  &  444.5   & 363.3*  & 423.7  & 429.6  & 510.2  & 560.8  \\
d=200   & 1424.9 & 868.7  & 761.0  & 715.6  & 647.2  & 620.0  & 607.5*  & 641.1  & 754.1  & 921.9  & 942.1  \\
d=500   & 1848.4 & 1549.9 & 1281.7 & 1213.5 & 1392.9 & 1229.9* & 1301.8 & 1307.8 & 1466.4 & 1586.4 & 1624.8\\ 
\hline
\end{tabular}}
\label{tb:expInx_regret}
\end{table}

In general, how much exploration is needed is an interesting open question that is beyond the scope of this paper. Recent works from \cite{bastani2021mostly} and \cite{russo2017tutorial} have identified regimes when exploration is unnecessary. They explain beautifully why exploration is wasteful when context vectors (in our case, the feature vectors) carry enough randomness. For those cases, greedy algorithms like \cite{oh2020sparsity} and \cite{li2021simple} will be more appropriate.

	\section{Summary and concluding remarks}\label{sec:conclusion}

Online advertising presents a number of distinctive challenges, including (a) a need for active exploration to speed the search for profitable audience-ad combinations, (b) many combinations to choose from, giving rise to high-dimensional bandit formulations, and (c) very low success probabilities, typically just a fraction of one percent. To the best of our knowledge, the LRDL method proposed in this paper is the first algorithm that meets all of those challenges. In particular, by incorporating the debiased lasso estimator of \cite{javanmard2014confidence}, it is able to explore actively without the relative symmetry assumption that underlies the greedy-based algorithms of \cite{oh2020sparsity} and \cite{li2021simple}.

An essential feature of our method is its use of a regression model to promote cross-learning, with the inclusion of ``interaction terms'' (see Subsection \ref{2sec:prob}) to reduce the threat of model misspecification, regularized to ensure that the cross-learning objective remains paramount.  We use a generalized version of classical Thompson sampling for treatment selection, incorporating a hyper-parameter $\rho$ that controls the extent of exploration: $\rho=0$ corresponds to a greedy algorithm, and $ \rho=1$ recovers classical Thompson sampling.

Our numerical experiments illuminate several side issues that are significant in the context of online advertising. One of these concerns the value, from an advertiser’s perspective, of a platform conducting experiments on the advertiser’s behalf. A platform has additional information about the users for whom ads are displayed, beyond what is available to the advertiser, and further has the flexibility to base ad choice on that finer information. In principle, then, an advertiser should be better off when a platform conducts experiments on the advertiser’s behalf. In our simulation studies, this is indeed true when sequential decisions are made using the LRDL method, but perhaps surprisingly, using more detailed information may actually leave the advertiser \textit{worse off} if a naive bandit algorithm is deployed. Roughly speaking, this happens when a simplistic algorithm is overwhelmed by the apparent need to sort through a multitude of decision options.

Our numerical experiments with the adjustable exploration parameter $\rho$ are consistent with previous findings that classical Thompson sampling might over-explore, cf. \cite{russo2017tutorial}. In fact, we have found values as small as $\rho=0.1$ to be optimal in our experiments, with the optimal value decreasing as the problem dimension (that is, the number of audience-ad combinations available) increases. Given the popularity of Thompson sampling, both in practice and in theoretical studies, this is a subject worthy of further investigation. Another obvious goal in future research is to establish performance guarantees for the LRDL method, which will require a theoretical analysis of debiased lasso behavior with adaptive data.

\ACKNOWLEDGMENT{The authors thank Mohsen Bayati, Carlos Carrion, Tong Geng, Peter Glynn, Xiliang Lin, Haim Mendelson, Daniela Saban, Stefan Wager and seminar participants at JD-Silicon Valley, NYU Stern MOILS, INFORMS, RM\&P, for helpful comments; and Jack Lin and Paul Yan of JD.com for helpful collaboration and support.}

	\begin{APPENDICES}
		\section{Summary of notation}\label{app_notation}
 \begin{itemize}
    \item $t$ is the time index; $t= 1, 2, \cdots, T$ where $T$ is the total numbr of trials to be conducted.
    \item $\KK = \{1, 2, \cdots, k, \cdots, K\}$ is the set of target audiences (TA), pre-specified by the platform.
    \item $\RR = \{1, 2,\cdots, r, \cdots, R\}$ is the set of creatives provided by the advertiser.
    \item $y_{t} \in \{0, 1\}$: binary variable indicating click or no click at time $t.$
    
    \item Advertiser's problem (single-stage model):
    \begin{itemize}
        \item $p_{kr}$ is the click-through probability for TA $k$ when creative $r$ is displayed.
        \item In LRDL, we assume $p_{kr} = \left(1+\exp(-\phi(k,r)^\top \theta)\right)^{-1} $, where $\phi(k,r)\in \RR^{KR}$ is the feature vector for TA-creative pair ($k,r$), and $\theta \in  \RR^{KR}$ is the unknown parameter vector to learn. 
        \item $\pi^* = \max_{k,r} p_{kr} $ is the optimal reward in the basic model.
        \item $\pi_{t} = p_{k_t r_t} $ is the advertiser's expected reward at time $t$ by choosing the (target audience, creative) combination $(k_t, r_t).$
        \item $R_T = \EE\left( \sum_{t=1}^T (\pi^* - \pi_{t})\right)$: the expected cumulative regret (also called ex-ante regret) up to time $T.$
    \end{itemize}

    \item Platform's problem (two-stage model):
        \begin{itemize}
            \item $\JJ = \{1, 2, \cdots, j, \cdots, J\}$ is the set of disjoint audiences (DAs), generated by partitioning the target audiences into non-overlapping sub-populations.
            \item $\PP(j|k)$: conditional probability of DA $j$ when the user is from TA $k.$
            
            \item $p_{jr}$ is the click-through probability for DA $j$ given that creative $r$ is displayed. 

            \item In LRDL, we assume $p_{jr} =  \left(1+\exp(-\phi(j,r)^\top \theta)\right)^{-1}  $, where $\phi(j,r)\in \RR^{JR}$ is the feature vector for DA-creative pair $(j,r)$, and $\theta \in  \RR^{JR}$ is the unknown parameter vector to learn. 
 
            \item $\pi^{*} = \max_{k,f(\cdot)} \sum_j \PP(j|k)~ p_{j~f(j)} $.
            
            \item $\pi_{t} = \sum_j \PP(j|k)  p_{j~f_t(j)} $ is the advertiser's expected reward at time $t$ by choosing target audience group $k_t$ and specifying a policy $f_t(\cdot)$ that maps the arriving DA to creative $f_t(\cdot): \JJ \to \RR $.

            \item $R_T = \EE \left(\sum_{t=1}^T (\pi^{*'} - \sum_j \PP(j|k_t)~p_{j~f_t(j)})\right)$: the expected cumulative regret up to time $T.$
        \end{itemize}
\end{itemize}
\section{Advertiser's problem vs. Platform's problem}\label{app_scenarios}

\begin{table}[H]   
        \centering
        \begin{tabular}{|| c|| c |c ||}
         \hline
          & Advertiser's problem (single-stage)  & Platform's problem (two-stage) \\ 
         \hline
        Information  &   $y_t$& $j_t, y_t$ (ex-ante)\\ [0.5ex] 
         \hline  
        Control & $(k_t, r_t)$ &   $k_t, f_t: DA \rightarrow CA$  \\ 
         \hline
         Objective & $\max \sum_{t=1}^T \sum_j \mathbb{P}(j|k_t)~ p_{j r_t} $  & $  \max \sum_{t=1}^T \sum_j \mathbb{P}(j|k_t)~ p_{ j f_t(j)} $\\
         \hline
         Regret & $T\cdot \pi^* -\sum_{t=1}^T \sum_j \mathbb{P}(j|k_t)~ p_{j r_t} $ & $T\cdot \pi^{*} -\sum_{t=1}^T \sum_j \mathbb{P}(j|k_t)~ p_{j f_t(j) } $\\
         \hline
        Effective dim & $K\cdot R$ & $J\cdot R \sim 2^K \cdot R $ \\
     \hline \hline
        LRDL  &$p_{kr} = \left(1+ \exp(-\theta^\top \phi(k,r))\right)^{-1} $&$p_{jr} = \left(1+ \exp(-\theta^\top \phi(j,r))\right)^{-1}$\\
        \hline
        SA Lasso & $p_{kr} = \left(1+ \exp(-\theta^\top \phi(k,r))\right)^{-1} $ & \NA \\ 
        DBBM & $p_{jr} \sim \text{Beta}(\alpha_{kr}, \beta_{kr})$&$p_{kr} \sim \text{Beta}(\alpha_{jr}, \beta_{jr})$\\ 
        \hline
        GLM UCB & $p_{kr} = \left(1+ \exp(-\theta^\top \phi(k,r))\right)^{-1} $ & \NA \\ 
        \hline
        LMLA & $p_{kr} = \left(1+ \exp(-\theta^\top \phi(k,r))\right)^{-1} $  & \NA \\ 
        \hline
        \end{tabular} 

    \end{table}
		
\section{Construction of the test problem based on real data}\label{app:real}

The data from \cite{geng2020comparison} consists of 933 experiments that were conducted on behalf of various advertisers by the JD.com platform, in 2019 and 2020. In each experiment, an advertiser provided a set of creatives (CAs) and specified a set of the target audiences (TAs), each drawn from the larger collection of TAs for which the platform operated an impressions market. The platform would then include in its experiment any site visitor belonging to one of the specified TAs, and would display to that visitor one of the CAs provided by the advertiser, using a selection algorithm (more specifically, a bandit algorithm) whose details need not concern us here. Some experiments were terminated by the advertiser and others by the platform.

To estimate the true CTR for a given TA-CA combination, one intuitive approach is to use 
 \eqs{\label{eq:CTR_est}\text{(number of clicks observed)}/\text{(number of impressions served)}}
for that specific combination. Unfortunately, such estimates are often inaccurate, because the adaptive sampling used in bandit experiments leads to many TA-CA combinations having too few trials to generate accurate CTR estimates. Also, the individual experiments conducted by \cite{geng2020comparison} were all too small, typically involving fewer than 10 TA-CA combinations, to provide a meaningful test of our LRDL method. Thus it is necessary to aggregate data from their many experiments in constructing our test problem.

To start, we determined the smallest possible set of disjoint audience segments (DAs) such that each of the TAs involved in any of the 933 experiments can be expressed as a finite union of such DAs. In this sense, the DAs we defined constitute a \textit{minimal partition} of site visitors involved in the experiments.  Record keeping for the experiments was detailed enough that we could associate a DA-CA pair with each site visitor involved in the experiments, not just a TA-CA pair. 

Next, we estimated a CTR for each DA-CA combination via \eqref{eq:CTR_est}, and eliminated from further consideration those combinations for which either the estimated CTR was $>$10\%, and therefore \textit{a priori} implausible, or else fewer than 500 impressions had been served. The remaining DA-CA combinations in the data set will hereafter be referred to as \textit{surviving} combinations. We used the Apriori algorithm of \cite{aggarwal2014frequent} to determine a maximal set of $J$ DAs and $R$ CAs such that each of the $J\times R$ combinations is a surviving combination; these will be referred to as the sets of surviving DAs and surviving CAs, respectively.  In summary, we have a reasonable CTR estimate for each combination of a surviving DA and surviving CA, but if any DA or CA were added to the sets we have identified, there would be at least one DA-CA combination for which a reasonable CTR estimate is lacking. Finally, we created a maximal set of $K$ TAs involved in the experiments such that each of them can be expressed as a finite union of surviving DAs.

Using this method, we arrived at a test problem with $R=26$ CAs and $J= 11$ DAs, corresponding to $K=4$ TAs. The conditional probability of DA $j$ given TA $k$ (see Subsection \ref{subsec:2stagealg}) is given by the obvious relationship
\[ \PP(j|k) = \frac{\text{number of impressions from DA $j$}}{\text{number of impressions from TA $k$}}~\text{if DA $j$} \subset \text{TA $k$},\]
and the CTR for each TA-CA combination is 
\[p_{kr} = \sum_{j} \PP(j|k) p_{jr}.\]

Overall, the estimated $p_{kr}$ values have mean $1.9\times 10^{-3} $ and standard deviation $2.26\times 10^{-3} $, with a maximum of $ 1.6 \%$ and a minimum of $0.019 \% $. For the empirical evaluation of our proposed method, we use the estimated CTRs as true CTRs. However, to ensure confidentiality, the CTR data displayed in the table is perturbed. Table \ref{tb:realCTR} shows the CTR table for the test problem of four TAs and 26 creatives. For the two-stage model developed in Section \ref{sec:twostage}, table \ref{tb:2stage_CTR} shows the  CTR table for the test problem of 11 DAs and 26 creatives. Table \ref{tb:TADA} displays the conditional probability of DA $j$ given TA $k$ for all pairs.

\begin{table}
\caption{CTR table for TA-CA pairs ($K =4, R = 26$)}
\smallskip
\centering
\begin{tabular}{l|llll} \hline
CTR  & TA1     & TA2     & TA3     & TA4     \\ \hline
CA1  & 0.00126 & 0.00109 & 0.00125 & 0.00236 \\
CA2  & 0.00171 & 0.00146 & 0.00175 & 0.00485 \\
CA3  & 0.00196 & 0.00185 & 0.00179 & 0.00202 \\
CA4  & 0.00196 & 0.00142 & 0.00195 & 0.00665 \\
CA5  & 0.00632 & 0.00496 & 0.00616 & 0.01449 \\
CA6  & 0.00058 & 0.00059 & 0.00058 & 0.00035 \\
CA7  & 0.00209 & 0.00089 & 0.00207 & 0.01612 \\
CA8  & 0.00111 & 0.00114 & 0.00114 & 0.00124 \\
CA9  & 0.00098 & 0.00064 & 0.00098 & 0.00314 \\
CA10 & 0.00084 & 0.00090 & 0.00084 & 0.00066 \\
CA11 & 0.00169 & 0.00158 & 0.00170 & 0.00324 \\
CA12 & 0.00156 & 0.00140 & 0.00153 & 0.00531 \\
CA13 & 0.00169 & 0.00196 & 0.00181 & 0.00120 \\
CA14 & 0.00071 & 0.00076 & 0.00078 & 0.00118 \\
CA15 & 0.00041 & 0.00042 & 0.00040 & 0.00020 \\
CA16 & 0.00082 & 0.00076 & 0.00081 & 0.00092 \\
CA17 & 0.00166 & 0.00159 & 0.00163 & 0.00281 \\
CA18 & 0.00185 & 0.00175 & 0.00188 & 0.00327 \\
CA19 & 0.00178 & 0.00175 & 0.00175 & 0.00370 \\
CA20 & 0.00083 & 0.00077 & 0.00085 & 0.00184 \\
CA21 & 0.00059 & 0.00060 & 0.00059 & 0.00054 \\
CA22 & 0.00131 & 0.00140 & 0.00140 & 0.00099 \\
CA23 & 0.00090 & 0.00132 & 0.00111 & 0.00776 \\
CA24 & 0.00066 & 0.00062 & 0.00066 & 0.00054 \\
CA25 & 0.00106 & 0.00112 & 0.00112 & 0.00021 \\
CA26 & 0.00225 & 0.00217 & 0.00241 & 0.00418 \\ \hline
\end{tabular}
\label{tb:realCTR}
\end{table}

\begin{table}
\caption{CTR table for DA-CA pairs ($J=11$, $R = 26$)}
\smallskip
\centering
\begin{adjustwidth}{-1cm}{}
\begin{tabular}{l|lllllllllll} \hline
CTR  & DA1     & DA2     & DA3     & DA4     & DA5     & DA6     & DA7     & DA8     & DA9     & DA10    & DA11    \\ \hline
CA1  & 0.00124 & 0.00132 & 0.01238 & 0.01047 & 0.00000 & 0.00227 & 0.00107 & 0.00000 & 0.00000 & 0.01462 & 0.00363 \\
CA2  & 0.00166 & 0.00188 & 0.01460 & 0.00510 & 0.00000 & 0.00000 & 0.00144 & 0.00000 & 0.00000 & 0.00000 & 0.00989 \\
CA3  & 0.00198 & 0.00000 & 0.00000 & 0.00000 & 0.00000 & 0.01380 & 0.00198 & 0.00000 & 0.00000 & 0.00000 & 0.00000 \\
CA4  & 0.00309 & 0.00000 & 0.00000 & 0.00000 & 0.00000 & 0.00000 & 0.00151 & 0.00000 & 0.00000 & 0.00000 & 0.01599 \\
CA5  & 0.01413 & 0.00000 & 0.00000 & 0.02208 & 0.01747 & 0.01260 & 0.00523 & 0.00000 & 0.00000 & 0.02465 & 0.02845 \\
CA6  & 0.00000 & 0.00000 & 0.00000 & 0.00336 & 0.00248 & 0.00160 & 0.00062 & 0.00000 & 0.00000 & 0.00000 & 0.00000 \\
CA7  & 0.00198 & 0.00192 & 0.00906 & 0.00156 & 0.00196 & 0.00235 & 0.00084 & 0.00000 & 0.00000 & 0.00000 & 0.03591 \\
CA8  & 0.00000 & 0.00190 & 0.00000 & 0.00761 & 0.00197 & 0.00277 & 0.00115 & 0.00000 & 0.00000 & 0.00000 & 0.00000 \\
CA9  & 0.00939 & 0.00433 & 0.00000 & 0.00000 & 0.00000 & 0.00000 & 0.00053 & 0.00000 & 0.00000 & 0.00000 & 0.00347 \\
CA10 & 0.00000 & 0.00168 & 0.00000 & 0.00000 & 0.00000 & 0.00000 & 0.00090 & 0.00000 & 0.00000 & 0.00000 & 0.00000 \\
CA11 & 0.00294 & 0.00280 & 0.00416 & 0.00169 & 0.00000 & 0.00337 & 0.00152 & 0.00238 & 0.00232 & 0.00000 & 0.00396 \\
CA12 & 0.00000 & 0.00000 & 0.03352 & 0.00000 & 0.00000 & 0.00000 & 0.00119 & 0.02554 & 0.00334 & 0.00000 & 0.01277 \\
CA13 & 0.00000 & 0.00307 & 0.02790 & 0.00698 & 0.00000 & 0.00000 & 0.00181 & 0.00000 & 0.00318 & 0.00000 & 0.00000 \\
CA14 & 0.00000 & 0.00301 & 0.00000 & 0.00652 & 0.00000 & 0.00000 & 0.00070 & 0.00000 & 0.00000 & 0.00000 & 0.00000 \\
CA15 & 0.00051 & 0.00051 & 0.00000 & 0.00000 & 0.00000 & 0.00000 & 0.00043 & 0.00000 & 0.00000 & 0.00000 & 0.00000 \\
CA16 & 0.00186 & 0.00097 & 0.00000 & 0.00000 & 0.00000 & 0.00000 & 0.00078 & 0.00000 & 0.00000 & 0.00000 & 0.00130 \\
CA17 & 0.00247 & 0.00223 & 0.00509 & 0.00499 & 0.00412 & 0.00732 & 0.00152 & 0.00084 & 0.00301 & 0.00383 & 0.00162 \\
CA18 & 0.00404 & 0.00344 & 0.00590 & 0.00469 & 0.00290 & 0.00285 & 0.00165 & 0.00230 & 0.00274 & 0.00285 & 0.00330 \\
CA19 & 0.00000 & 0.00323 & 0.00477 & 0.00383 & 0.00487 & 0.00619 & 0.00167 & 0.00217 & 0.00194 & 0.00133 & 0.00314 \\
CA20 & 0.00230 & 0.00177 & 0.00884 & 0.00432 & 0.00359 & 0.00098 & 0.00067 & 0.00000 & 0.00164 & 0.00312 & 0.00201 \\
CA21 & 0.00075 & 0.00138 & 0.00000 & 0.00000 & 0.00000 & 0.00000 & 0.00060 & 0.00000 & 0.00000 & 0.00000 & 0.00000 \\
CA22 & 0.00117 & 0.00077 & 0.00627 & 0.00204 & 0.00000 & 0.00000 & 0.00136 & 0.00000 & 0.00393 & 0.00000 & 0.00165 \\
CA23 & 0.00000 & 0.01540 & 0.07591 & 0.00000 & 0.00000 & 0.00000 & 0.00031 & 0.00000 & 0.01155 & 0.00000 & 0.00417 \\
CA24 & 0.00093 & 0.00000 & 0.00000 & 0.00000 & 0.00000 & 0.00000 & 0.00067 & 0.00000 & 0.00000 & 0.00000 & 0.00130 \\
CA25 & 0.00000 & 0.00053 & 0.00000 & 0.00643 & 0.00000 & 0.00000 & 0.00118 & 0.00000 & 0.00000 & 0.00000 & 0.00000 \\
CA26 & 0.00298 & 0.00250 & 0.00667 & 0.00818 & 0.00000 & 0.00000 & 0.00205 & 0.00621 & 0.00616 & 0.00000 & 0.00770 \\ \hline
\end{tabular}
\label{tb:2stage_CTR}
\end{adjustwidth}

\end{table}
\begin{table}[]

\caption{Conditional probability $\PP(j|k)$ for each DA $j$ and TA $k$. }
\smallskip
\centering
\begin{tabular}{l|llll} \hline
$\PP(j|k)$ & TA1    & TA2    & TA3    & TA4    \\ \hline
DA1  & 0.0279 & 0      & 0.0278 & 0      \\
DA2  & 0.0315 & 0.0333 & 0.0313 & 0.3915 \\
DA3  & 0      & 0.0038 & 0      & 0      \\
DA4  & 0      & 0      & 0.0100 & 0      \\
DA5  & 0.0037 & 0.0040 & 0      & 0.0466 \\
DA6  & 0.0117 & 0      & 0      & 0.1461 \\
DA7  & 0.8823 & 0.9354 & 0.8789 & 0      \\
DA8  & 0.0035 & 0.0037 & 0      & 0      \\
DA9  & 0      & 0.0199 & 0.0187 & 0      \\
DA10 & 0.0060 & 0      & 0      & 0      \\
DA11 & 0.0334 & 0      & 0.0333 & 0.4158 \\ \hline
\end{tabular}
\label{tb:TADA}
\end{table}

		\section{Calculation of Algorithms \ref{alg_LRDL}} \label{app:calc}

To start, we compute the original Lasso estimator. Let $\cL_{\tau}( \theta)$ be the normalized negative log-likelihood corresponding to the observation $( y_{\tau}, x_\tau)$ at time $\tau$, where $x_\tau = \phi(k_\tau, r_\tau)$:
\eqss{\cL_{\tau}( \theta) & =  -\log f(y_{\tau}| x_\tau)\\
& = -y_\tau \theta^\top  x_\tau + \log(1+ \exp(\theta^\top x_\tau))
}
Define the loss function:
\eqss{\cL(\theta ) & = \frac{1}{t} \sum_{\tau=1}^t \cL_{\tau}(\theta) \\
& =   \frac{1}{t} \sum_{\tau=1}^t [  -y_\tau \theta^\top  x_\tau + \log(1+ \exp(\theta^\top x_\tau))] }
The lasso estimator minimizes the $\ell_1$ penalized loss function:
\eqss{
\hat{\theta}_l & =  \frac{1}{t} \sum_{\tau=1}^t [  -y_\tau \theta^\top  x_\tau + \log(1+ \exp(\theta^\top  x_\tau))]  + \lambda ||\theta||_1
}
To compute the debiasing term, we first compute the Fisher information matrix of $f(y| x_\tau)$,
\eqss{
\cI_{\tau}(\hat{\theta}_l) & = - \EE( \nabla^2 \log f (y| x_\tau) \\
& = (1+\exp(\hat{\theta}_l^\top x_\tau))^{-1} (1+\exp(-\hat{\theta}_l^\top x_\tau))^{-1} x_\tau^\top x_\tau
}
Hence, the sample covariance matrix $\hat{\Sigma}(\hat{\theta}_l)$ is: 
\eqs{ \label{eq:debiasVar}
\hat{\Sigma} & = \frac{1}{t} \sum_{\tau=1}^t \cI_{\tau}(\hat{\theta}_l) \\
& =\frac{1}{t} \sum_{\tau=1}^t\left[ (1+\exp(\hat{\theta}_l^\top x_\tau))^{-1} (1+\exp(-\hat{\theta}_l^\top x_\tau))^{-1} x_\tau^\top x_\tau\right]
}
When $ \hat{\Sigma}$ is invertible ($t\geq RK$), $M =\hat{\Sigma}^{-1} $. When $ \hat{\Sigma}$ is singular($t< RK$), we solve 
 \[ m_i = \argmin_m ~ m^\top \hat{\Sigma} m 
 \text{ s.t.} ||\hat{\Sigma} m - e_i||_{\infty } \leq \mu \]
where $e_i \in \RR_d$ is the vector with one at the $i$-th position and zero elsewhere. Set  $M:= [m_1, \cdots, m_d]^\top$. If any of the above problems is not feasible, set $M = I_{d \times d}$. The debiased Lasso estimator $\hat{\theta}_d$ is
\eqs{\label{eq:debias}
  \hat{\theta}_d  &= \hat{\theta}_l + \frac{1}{t} M \sum_{\tau=1}^t (y_\tau  (1+\exp(-\hat{\theta}_l^\top x_\tau))^{-1} )x_\tau }
  and the covariance matrix 
  \eqs{\label{eq:debiasVar}  \Sigma_d & = M^\top\hat{\Sigma} M / t.
}
In summary, at time  $ t$, $\theta \sim \cN( \hat{\theta}_d ,   \Sigma_d )$, where $ \hat{\theta}_d$ and $ \Sigma_d  $ are given in equations \eqref{eq:debias} and  \eqref{eq:debiasVar}.

	\end{APPENDICES}
	
	\bibliographystyle{ormsv080}
	\bibliography{reference}
	
\end{document}